\documentclass{article}

\usepackage[preprint]{neurips_2026}

\usepackage[utf8]{inputenc}
\usepackage[T1]{fontenc}
\usepackage{hyperref}
\usepackage{url}
\usepackage{booktabs}
\usepackage{amsfonts}
\usepackage{verbatim}
\usepackage{enumitem}
\usepackage{nicefrac}
\usepackage{microtype}
\usepackage{xcolor}
\usepackage{amsmath}
\usepackage{amsthm}
\usepackage{mathtools}

\newtheorem{theorem}{Theorem}

\usepackage{graphicx}
\usepackage{multirow}
\usepackage{booktabs}
\usepackage{subcaption}
\usepackage{longtable}

\newtheorem{definition}{Definition}

\title{Neural Expansion: A Unified Mechanism for How Deep Neural Networks Generalize}

\author{\textbf{Chashi Mahiul Islam}$^{1}$ \quad
  \textbf{Samuel Jacob Chacko}$^{1}$ \quad
  \textbf{Mao Nishino}$^{2}$ \\
  \textbf{Canlin Zhang}$^{3}$ \quad
  \textbf{Xiuwen Liu}$^{1}$ \\
  $^{1}$Department of Computer Science, Florida State University \\
  $^{2}$Department of Mathematics, Florida State University \\
  $^{3}$Independent \\
  \texttt{\{cislam,sjacobchacko,mnishino,xliu\}@fsu.edu} \\
  \texttt{canlingrad@gmail.com}
}

\begin{document}

\maketitle

\begin{abstract}
Deep neural networks demonstrate remarkable generalization despite being highly over-parameterized, challenging traditional statistical learning theories. While several theoretical paradigms such as margin-based bounds, PAC-Bayesian analyses, and neural tangent kernels provide global explanations, they often lack mechanistic insight into how generalization emerges from training dynamics. We introduce Neural Expansion, a unified geometric framework that characterizes generalization through the expansion of directionally-selective ray coverage in the input space. Using Generalization Intervals (GIs), we quantify direction-specific regions around training samples where the model's predictions remain stable. Our extensive empirical studies across diverse architectures and datasets reveal that training leads to progressive expansion of ray coverage, particularly along directions with low Jacobian singular values. This mechanism helps explain a large fraction of correctly classified test inputs, adversarial examples, and even mislabeled or out-of-distribution samples through the geometry of ray coverage rather than instance-level memorization. By connecting input-space curvature to predictive behavior, our framework provides a mechanistic foundation that complements existing global theories of deep learning. Code and instructions for running the experiments are available at \url{https://github.com/meetchashi/neural_expansion_neurips_2026}.
\end{abstract}

\section{Introduction}
The remarkable generalization capability of deep neural networks presents a fundamental theoretical challenge. Despite having millions or billions of parameters, far exceeding their training dataset size, these models generalize well to unseen examples, contradicting classical learning theory predictions \citep{zhang2016understanding}. This generalization persists even without explicit regularization, suggesting intrinsic mechanisms beyond conventional capacity control \citep{neyshabur2017pac, bartlett2017spectrally}.

Traditional frameworks for understanding generalization, such as VC-dimension theory and uniform convergence bounds, provide vacuous guarantees for modern neural networks. While recent approaches, including spectral norm-based margins \citep{bartlett2017spectrally}, PAC-Bayesian bounds \citep{neyshabur2017pac}, compression-based complexity \citep{arora2018stronger}, neural tangent kernels \citep{jacot2018neural}, simplicity bias \citep{valle2018deep}, sharpness-aware training \citep{andriushchenko2022sam, andriushchenko2023lowrank}, and input-space geometric invariance \citep{movahedi2025geometric}, have advanced our understanding, a critical question remains unaddressed: why does a specific test input receive a specific classification?

We propose \textbf{Neural Expansion}, a novel geometric framework explaining how networks develop directionally-selective ray coverage around training samples during learning. Our approach is based on \textit{Generalization Intervals (GIs)}, which quantify directionally-sensitive regions around training points within which a model's classification remains invariant~\citep{salman2020towards}. Rather than defining complex decision boundaries in high-dimensional spaces, we analyze how networks progressively expand stable classification zones, \textit{ray coverage}, in specific directions.

Neural Expansion refers to the progressive growth of this ray coverage during training, which becomes more pronounced in directions with low Jacobian singular values. Unlike margin-based approaches that focus on decision boundary distances, Neural Expansion captures direction-specific invariance properties that explain why certain test points get assigned to particular classes. Furthermore, while kernel theories (NTK) assume a fixed feature space and neural collapse focuses on terminal phases, our framework explains the dynamic formation of ray coverage throughout the entire training process.

Our key contributions include:
\begin{itemize}
    \item A theoretical framework that explains sample-level generalization through directional stability in input space, connecting Jacobian singular values to classification behavior

    \item Empirical demonstration that neural networks form wide generalization intervals in directions with low curvature, creating expansive, overlapping ray coverage around training samples

    \item Evidence that test-time classifications occur because test samples fall within the ray coverage of training samples, not because of instance-level memorization

    \item A geometric perspective showing how standard generalization, adversarial vulnerability, long-tailed misclassification, and random-label learning can be interpreted through the directional structure of generalization intervals
\end{itemize}

In this work, we establish that the local, sample-level geometry of neural network behavior is not merely an empirical curiosity but a theoretically grounded consequence of gradient-descent training. This perspective is complementary to existing theories of generalization, which provide valuable insights into optimization dynamics and global complexity measures. By formally analyzing how Jacobian structure creates and expands directional corridors of stability in input space, we provide a missing link between representation geometry and classification decisions, one that applies across diverse architectures and dataset distributions.

\section{Background and Related Work}
Understanding why overparameterized neural networks generalize well remains one of the central open questions in deep learning theory \citep{zhang2016understanding}. \citet{bengio2013representation} hypothesized that effective generalization emerges when learned representations disentangle underlying factors of variation. The success of deep networks has since spurred diverse theoretical perspectives. Our work introduces a local, mechanistic framework based on Generalization Intervals (GIs)~\citep{salman2020towards}, focusing on individual sample-level classification behavior through Jacobian-derived directional invariance.

Classical approaches rely on complexity-based measures such as VC dimension, Rademacher complexity, and margins. Spectral norm-based bounds \citep{bartlett2017spectrally}, PAC-Bayesian frameworks \citep{neyshabur2017pac}, and compression-based approaches \citep{arora2018stronger} offer compelling global insights. Our GI framework provides a local geometric lens, explaining generalization through stable directional regions around training samples where predictions remain invariant.

The Neural Tangent Kernel (NTK) approximates wide networks as kernel regression around initialization \citep{jacot2018neural}, formalized and contrasted with richer feature-learning regimes \citep{chizat2019lazy,woodworth2020kernel} and extended to explain triple descent \citep{adlam2020neural}. These approaches focus on global dynamics near initialization; our work instead quantifies post-training directional stability via the Jacobian, capturing piecewise-linear generalization regions in realistic regimes.

Neural Collapse describes how class means, features, and classifier weights align near convergence \citep{papyan2020prevalence,han2021neural,wu2024linguistic}, with recent work linking this to the average gradient outer product \citep{beaglehole2024average}. Our GI framework shows that classification decisions arise from stable, low-curvature regions that generalize across architectures and tasks.

Generalization has also been attributed to a simplicity bias \citep{valle2018deep}, supported by double and deep double descent \citep{belkin1812reconciling,zhang2016understanding,nakkiran2021deep}. Recent work shows that seeking flat loss regions improves generalization \citep{foret2020sharpness}, SGD implicitly minimizes complexity-promoting measures \citep{vardi2023implicit}, and networks can benignly overfit noisy data under favorable signal-to-noise conditions \citep{frei2023benign}, phenomena our framework explains geometrically via Jacobian curvature.

\section{A Theoretical Framework for Neural Network Generalization}
In this section, we develop our theoretical framework that captures the directional properties of generalization in deep neural networks. We introduce the \textbf{generalization interval (GI)}, which captures how far we can go in a given direction without changing the class, and connect it to Jacobian singular value analysis.

\begin{definition}
    Let $f : \mathbb{R}^d \to \mathbb{R}^c$ be a $c$-class classifier outputting the logits for each class, $x_0 \in \mathbb{R}^d$ be a given input data, $y\in \{1,\cdots ,c\}$ be its class and $d\in \mathbb{R}^d$ be the direction to traverse. The \textbf{generalization interval (GI)} of $f$ from the input $x_0$ in the direction of $d$ is defined by
    \begin{align}
        \textrm{GI}(f, x_0,d)  = \sup\{a>0| f^y(x_0+td)>\textnormal{max}_{i\neq y}f^i(x_0+td) \textrm{ for all }t\in[0,a]\}
    \end{align}
\end{definition}
As discussed earlier, this quantity measures how far we can go in the given direction before we reach the decision boundary. The idea is similar to the classical concept of \textbf{margins}, but a GI is direction-dependent: a model might exhibit large GIs in some directions and small GIs in others around the same point. Nevertheless, there is an important relationship between the margins and the GIs, which we will discuss later. The act of finding a GI can be described as shooting a \textit{ray} from the given input until we reach the decision boundary. As such, it is natural to define the following:
\begin{definition}
The \textit{ray coverage} of $x_0$ is the set of all points that lie within their respective GI bounds:
$$\text{RC}(x_0) = \left\{x : \text{GI}\left(f,x_0,\frac{x-x_0}{\|x-x_0\|}\right) \geq \|x-x_0\|\right\}$$
\end{definition}

\subsection{Jacobian Analysis and GI Formation}
\label{sec:jacobian-analysis}

To understand why neural networks exhibit directionality in ray coverages through sensitivity, we analyze the Jacobian of the network's representation function. Let $\phi: \mathbb{R}^d \to \mathbb{R}^m$ denote the mapping from input to the penultimate layer (i.e., the representation), and $W \in \mathbb{R}^{c \times m}$ be the final linear classifier, such that $f(\mathbf{x}) = W\phi(\mathbf{x})$.

The Jacobian $J_\phi(\mathbf{x}) \in \mathbb{R}^{m \times d}$ captures how the representation changes with respect to small perturbations in the input. We perform singular value decomposition:

\begin{equation}
J_\phi(\mathbf{x}) = U\Sigma V^T = \sum_{i=1}^{\min(m,d)} \sigma_i \mathbf{u}_i \mathbf{v}_i^T
\end{equation}

where $\sigma_1 \geq \sigma_2 \geq \cdots \geq \sigma_{\min(m,d)} \geq 0$ are the singular values, $\mathbf{v}_i \in \mathbb{R}^d$ are the right singular vectors (directions in input space), and $\mathbf{u}_i \in \mathbb{R}^m$ are the left singular vectors (directions in representation space).

Our empirical analysis reveals a crucial property: we observe that, in trained neural networks, the vast majority of singular values are close to zero (See Figure \ref{fig:subfig_b}). For instance, in a network trained on CIFAR-10 (3,072-dimensional input), typically only 10-15 singular values are significantly non-zero, while over 99.5\% of the singular value spectrum is effectively zero (Appendix~\ref{sec:jacobian-svd}). This spectral structure has profound implications for generalization due to Theorem \ref{thm: lin-appx-gi}. For a right singular vector direction $v$ of $J_\phi$, $\|J_f(x_0)v\|$ is proportional to the corresponding singular value. Therefore, the theorem predicts that the GI is large for all but a few limited directions due to the directional sensitivity of the model, which is consistent with the empirical observation we will present later in Figure \ref{fig:subfig_a}.

\subsection{Linear Approximation of the GI and the Relation to the Margin}
As the GI can be described as the margins of samples with directionality, they naturally have a close relationship. To describe it, we first define the margin:
\begin{definition}
    Let $f:\mathbb{R}^d\to \mathbb{R}^c$ be a $c$-class classifier, $x_0\in \mathbb{R}^d$ be a given input data, and $y\in \{1,\cdots , c\}$ be its class. The \textbf{margin} of the sample $(x_0, y)$ is $f^y(x_0) - \textnormal{max}_{i\neq y}f^i(x_0)$ and we denote it by $q_0$. Here, $f^i(x_0)\in \mathbb{R}$ means the $i$th component of $f(x_0)$.
\end{definition}
    We now have the following relation:
\begin{theorem}
    \label{thm: lin-appx-gi}
    Let $x_0, d\in \mathbb{R}^d$ be a given input and a given direction. Suppose $y\in \{1,\cdots c\}$ is the class of $x_0$. Assume that the classifier $f:\mathbb{R}^d\to \mathbb{R}^c$ is differentiable with an inhomogeneous directional derivative: the $y$th component of $J_f(x_0)d$ is not equal to any other component.  Consider the linear approximation $f(x_0 + \epsilon d) = f(x_0)+\epsilon J_{f}(x_0)d + R(x_0,\epsilon,d)$ where $R$ is the error term satisfying $\|R(x_0,\epsilon,d)\|_1<\delta$ for some $\delta>0$, $J_f(x_0)$ denotes the Jacobian matrix for $f$ at $x_0$ and $\|\cdot\|_1$ is the $\ell^1$-norm. Then
    \begin{equation}
        \frac{q_0 - \delta}{\|J_f(x_0)d\|_1 }\leq \textnormal{GI}(f,x_0, d)
    \end{equation}
\end{theorem}
The proof can be found in Appendix~\ref{sec: proof-thm}. This result tells us that, under linear approximation, a larger margin would indicate a larger generalization interval. This partially explains why the GIs should increase under gradient descent training, as previous literature such as \citep{Lyu2020Gradient} shows that a smooth approximation of the margin for a homogeneous network would monotonically increase under gradient descent. We note that, as we show later, the linear approximation holds surprisingly well (see Section \ref{sec: lin-apx}). Nonlinear effects are accounted for through the assumed residual bound $\delta$ (Appendix~\ref{app:taylor_remark}).

However, this is not the end of the story. The lower bound has the norm of the directional derivative $\|J_f(x_0)d\|$, and thus, sensitivity in each direction would contribute to the size of the GIs.

\subsection{GI Strictly Refines Spectral Margin Along Singular Directions}
Theorem~\ref{thm: lin-appx-gi} establishes that the GI lower bound grows with margin and shrinks with directional sensitivity. This raises a natural question: does GI offer strictly more information than classical spectral margin theory~\citep{bartlett2017spectrally}, or does it merely restate the same global bound in geometric language? We show that GI strictly refines spectral margin by decomposing input space direction-by-direction in a way that spectral margin cannot.

\begin{theorem}[GI--Spectral Margin Decomposition]
\label{thm:gi_spectral}
Under the same setting as Theorem~\ref{thm: lin-appx-gi}, let
$J_\phi(x_0) = U\Sigma V^\top$ be the SVD of the penultimate-layer Jacobian
with singular values $\sigma_1 \geq \cdots \geq \sigma_{\min(m,d)} \geq 0$
and right singular vectors $v_i \in \mathbb{R}^d$. Then for each $v_i$:
\begin{equation}
    \mathrm{GI}(f,\, x_0,\, v_i) \;\geq\;
    \frac{q_0 - \delta}{\sqrt{c}\,\|W\|_{\mathrm{op}}\,\sigma_i}
    \label{eq:gi_spectral_bound}
\end{equation}
As $\sigma_i \to 0$ with $q_0>\delta$, this bound grows without bound, while the spectral
margin bound remains fixed and direction-agnostic.
\end{theorem}

The bound decomposes input space into \emph{sensitive directions}
(large $\sigma_i$) and \emph{invariant directions}
(small $\sigma_i$). Spectral margin theory
captures only the former; GI captures both. Since a significant number of singular
directions in trained networks have $\sigma_i \approx 0$
(Section~\ref{sec:jacobian-analysis}), GI provides nontrivial directional
information in precisely the regime where spectral margin is silent.
The proof is in Appendix~\ref{app:proof_thm_gi_spectral}.

\subsection{Jacobian control of the input-loss gradient}
Theorem~\ref{thm:gi_spectral} reveals that directions with large singular values yield a small lower bound on GIs. A deeper question follows: do these high-sensitivity, low-GI directions correspond to the directions where the loss landscape is sharpest? To partially answer this question, we bound the input-loss gradient in terms of the largest Jacobian singular value, connecting the loss landscape and the most sensitive direction.

\begin{theorem}[GI and Loss Landscape Sharpness]
\label{thm:gi_sharpness}
Under the same setting as Theorem~\ref{thm: lin-appx-gi}, let
$\mathcal{L}(x) = -\log \sigma_y(f(x))$ be the cross-entropy loss at $x_0$
with true class $y$, and let $\nabla_x \mathcal{L}(x_0)$ be its gradient with
respect to the input. Then:
\begin{equation}
    \|\nabla_x \mathcal{L}(x_0)\|_2
    \;\leq\; \sqrt{2}\,\|W\|_{\mathrm{op}} \cdot \sigma_1
    \label{eq:loss_grad_bound}
\end{equation}
where $\sigma_1$ is the largest singular value of $J_\phi(x_0)$.
\end{theorem}

Equation~\eqref{eq:loss_grad_bound} shows that input-space sharpness is controlled by the top Jacobian singular value $\sigma_1$, the same quantity that governs GI via Theorem~\ref{thm:gi_spectral}. Empirically, the largest right singular vector $v_1$ also closely approximates the unconstrained steepest ascent direction, as shown in Figure~\ref{fig:adv_projection}. This connects Neural Expansion directly to the flat minima literature~\citep{keskar2017largebatchtrainingdeeplearning}: low-curvature directions (large index $i$, small $\sigma_i$) correspond to flat directions in the loss landscape and large GIs, while high-curvature directions (small index $i$, large $\sigma_i$) correspond to sharp directions and small GIs. The proof is in Appendix~\ref{app:proof_thm_gi_sharpness}.

\subsection{GI Quantifies the Adversarial Robustness Radius}
The GI provides a direct lower bound on the size of adversarial perturbations. We now show that every successful class-changing perturbation has norm at least the minimum GI over unit directions, and derive a further lower bound in terms of the margin and Jacobian norm.

\begin{theorem}[Adversarial Perturbation Radius via GI]
\label{thm:gi_adversarial}
Under the same setting as Theorem~\ref{thm: lin-appx-gi}, let
$\Delta \in \mathbb{R}^d$ be any perturbation that changes the
classification at $x_0$, i.e., $\arg\max_i f^i(x_0 + \Delta) \neq y$.
Then:
\begin{equation}
    \|\Delta\|_2 \;\geq\; \min_{\|d\|=1}\,\mathrm{GI}(f, x_0, d)
    \label{eq:adv_radius_gi}
\end{equation}
Furthermore, combining with Theorem~\ref{thm:gi_spectral}:
\begin{equation}
    \|\Delta\|_2 \;\geq\;
    \frac{q_0 - \delta}{\sqrt{c}\,\|W\|_{\mathrm{op}}\,\sigma_1}
    \label{eq:adv_radius_explicit}
\end{equation}
\end{theorem}

Equation~\eqref{eq:adv_radius_gi} establishes a bound on the adversarial robustness radius: large GIs in every direction guarantee robustness. Equation~\eqref{eq:adv_radius_explicit} makes this quantitative in terms of observable quantities: margin $q_0$, classifier norm $\|W\|_{\mathrm{op}}$, and the largest Jacobian singular value $\sigma_1$. The proof is in Appendix~\ref{app:proof_thm_gi_adversarial}.

\section{Experimental Setup}

Full implementation details are provided in Appendix~\ref{app:experimental_details}.

\textbf{Datasets:}
We evaluate on three benchmark datasets: MNIST~\citep{lecun1998gradient}, CIFAR-10~\citep{krizhevsky2009learning}, and ImageNette~\citep{deng2009imagenet}. To assess generalization under more challenging regimes, we include random label variants and long-tailed variants (following~\citet{zhang2016understanding} and~\citet{ma2022delving}). Full details on these datasets and their variants are provided in Appendix~\ref{app:datasets_and_variants}.

\textbf{Model Architectures:}
To demonstrate the universality of Neural Expansion, we evaluate three architecture families: Multi-Layer Perceptrons, Convolutional Networks (e.g., ResNet-20/50, VGG-16), and Vision Transformers (including ViT-B/16 and DeiT-S). All models are trained with SGD using dataset-appropriate hyperparameters until convergence. Full architecture and training details are provided in Appendix~\ref{app:model_architectures}.

\textbf{Generalization Interval Analysis:} Our primary analytical tool is the Generalization Interval (GI), which quantifies directional prediction stability in input space. GIs are computed using a step-wise search algorithm (Appendix~\ref{app:analysis-methods}) along RSVs of the penultimate-layer Jacobian (Section~\ref{sec:jacobian-analysis}), combined with ray coverage measurement toward training samples and linear approximation evaluation (Section~\ref{sec: lin-apx}). Together, these methods allow us to analyze the formation of ray coverage during training, its alignment with Jacobian structure, and its implications across standard, long-tailed, random-labeled, and adversarial test settings.

\section{Results}

\subsection{GI Behavior Across Model Types and Training Scenarios}
\label{sec:gi_evolution}
We conduct a systematic analysis of generalization interval (GI) behavior across model architectures, training scenarios, and datasets. GIs are computed along the right singular vectors (RSVs) of the input-to-penultimate-layer Jacobian. To ensure consistency, each RSV direction is scaled by a factor of 56.0, the maximum distance in the normalized 784-dimensional MNIST input space. Unless otherwise noted, all experiments use an MLP architecture (784--1024--512--10). Our results reveal consistent GI patterns across settings, pointing to a universal geometric structure underlying neural network generalization.

\subsubsection{Temporal Evolution of Generalization Intervals During Training}
\begin{figure}[ht!]
\centering
\includegraphics[width=\linewidth]{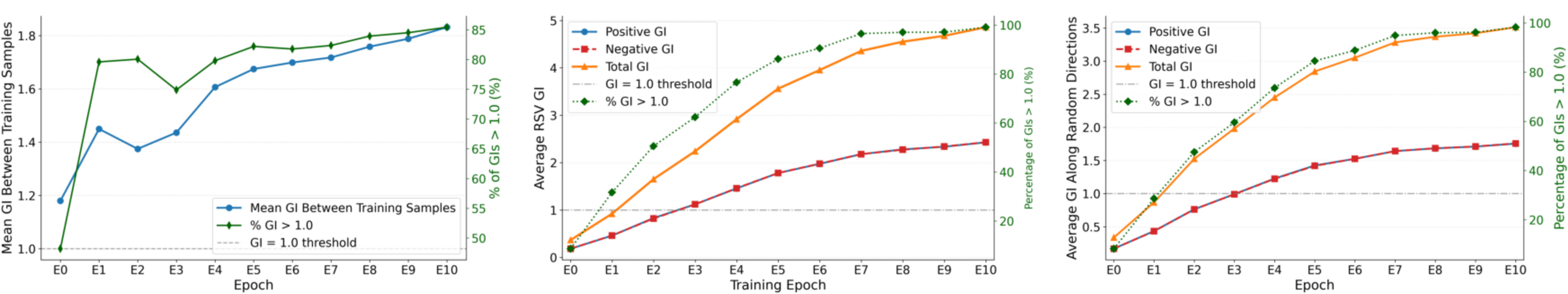}
\caption{
Evolution of ray coverage during training of an MNIST MLP over 10 epochs.
\textbf{Left:} Mean positive GI growth between same-class training samples, with percentage of GIs exceeding 1.0.
\textbf{Middle:} Average GI along RSV directions.
\textbf{Right:} Average GI along random directions.}
\label{fig:gi_evolution}
\end{figure}

We observe how GI-based ray coverage progressively develops during the training process in Figure~\ref{fig:gi_evolution}. Both the mean GI and the proportion exceeding 1.0 increase steadily, strongly correlating with improvements in test accuracy (Pearson $r=0.99$ across epochs, range 0.994--0.998 over three runs). While both RSV and random directions (isotropic Gaussians, unit-normalized) exhibit GI expansion, RSV directions consistently show larger GIs and faster growth rates, particularly in later epochs. This demonstrates the network's preferential expansion of ray coverage along low-curvature directions identified by the Jacobian SVD (Appendix~\ref{app:gi_evolution}).

\subsubsection{GI Expansion in Early vs. Later Training}
\label{sec: GI exp}
\begin{figure}[ht!]
    \centering
    \begin{subfigure}[b]{0.48\linewidth}
        \centering
        \includegraphics[width=\linewidth]{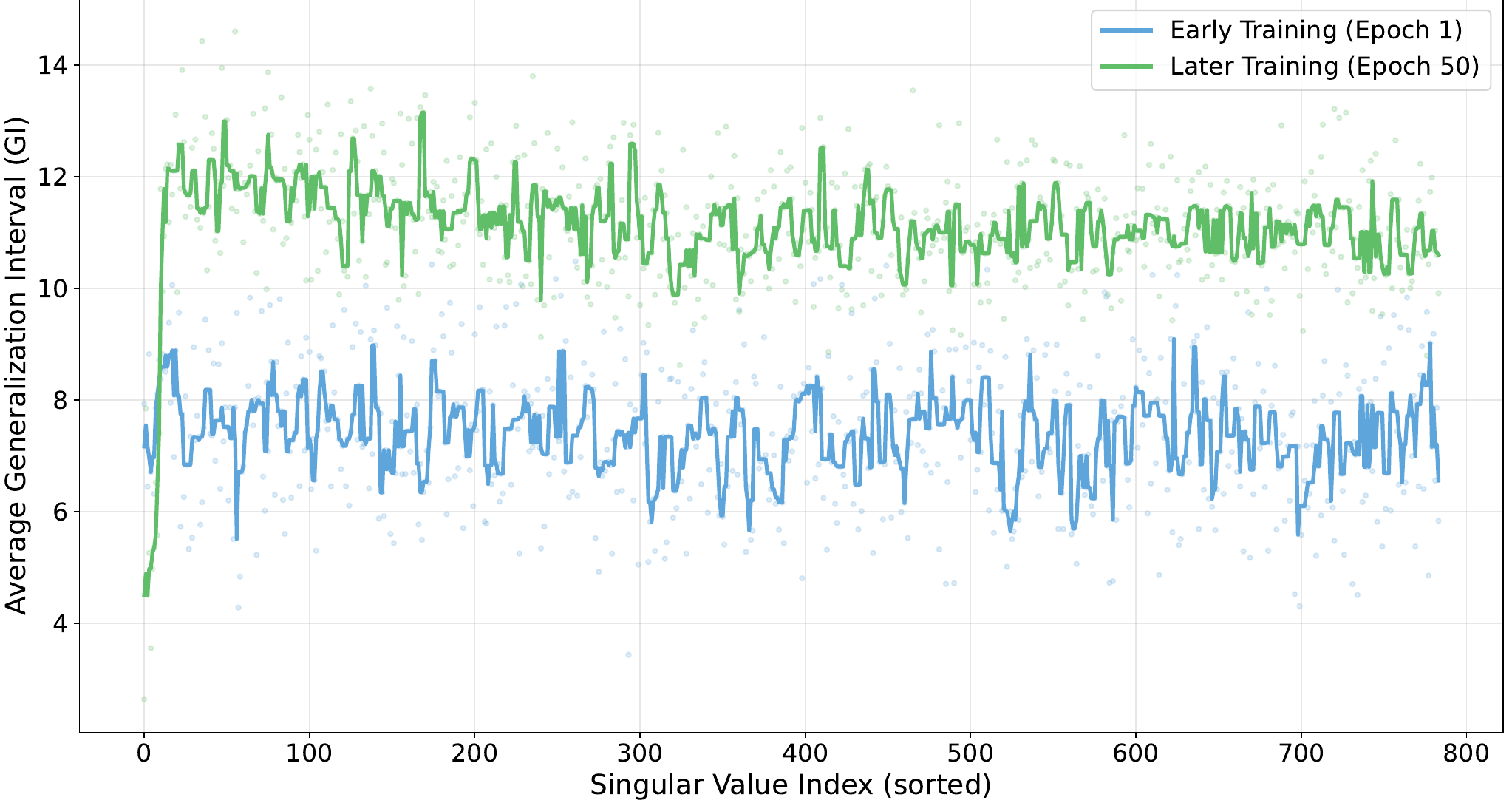}
        \caption{}
        \label{fig:subfig_a}
    \end{subfigure}
    \hfill
    \begin{subfigure}[b]{0.48\linewidth}
        \centering
        \includegraphics[width=\linewidth]{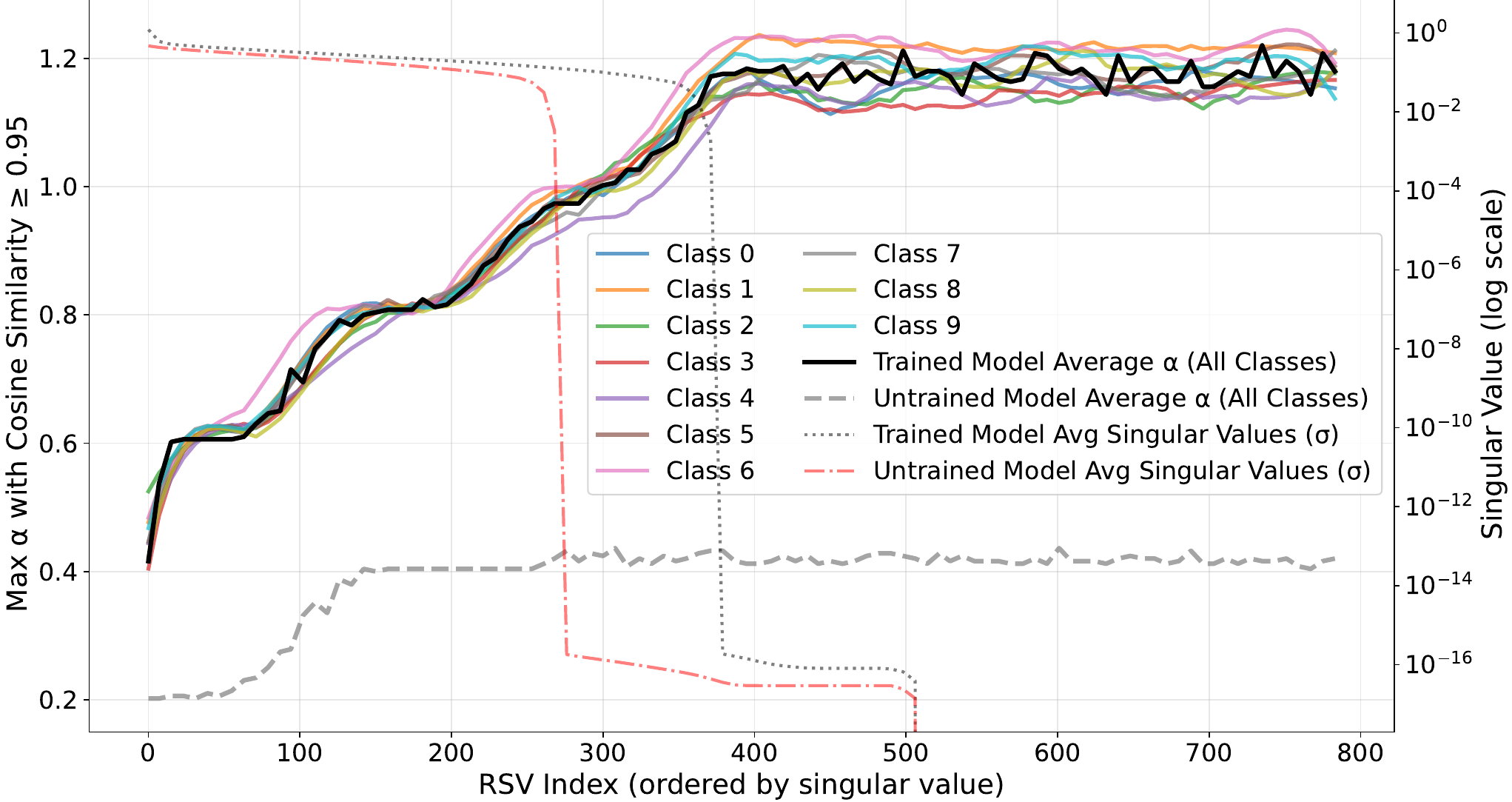}
        \caption{}
        \label{fig:subfig_b}
    \end{subfigure}
    \caption{(a) GI magnitudes across RSV directions in early- vs. late-stage models (epoch 1 vs. epoch 50). Training increases GIs selectively along high-index (low-curvature) directions. (b) Comparison of linear approximation quality along RSV directions before and after training.}
    \label{fig:gi_expansion_and_linear_approximation}
\end{figure}

Figure~\ref{fig:subfig_a} compares GI magnitudes across RSV directions between an early-stage model (epoch 1) and a fully trained model (epoch 50). In the early phase, GI values are relatively uniform, averaging around 7 across RSVs. After training, the distribution becomes skewed, GIs increase substantially along high-index (low-curvature) RSV directions and decrease along low-index (high-curvature) directions, with the average rising to 11.5. This indicates that training selectively expands ray coverage along directions associated with smaller singular values. We observe 512 non-zero singular values, corresponding to the representation layer dimension, while the remaining RSV directions are null due to zero singular values.

\subsubsection{Linear Approximation Along Directions with Different Singular Values}
\label{sec: lin-apx}

Figure~\ref{fig:subfig_b} quantifies the local linearity of the model along RSV directions by measuring the maximum distance $\alpha$ for which a linear approximation of the model output maintains a cosine similarity of at least 0.95 with the true output (evaluated in the penultimate-layer representation space).

In the untrained model (random initialization), linear approximation quality is uniformly low: $\alpha$ ranges from \textasciitilde0.2 in high-curvature directions (low RSV indices) to \textasciitilde0.4 in flatter directions, indicating limited local linearity across all directions. After training (49 epochs), the model exhibits a clear monotonic trend where linearity improves substantially with RSV index, starting from $\alpha \approx 0.4$ and reaching $\alpha \approx 1.2$ in low-curvature directions. This coincides with a slower decay in the Jacobian's singular values, which extend further into high-index RSVs compared to the untrained case.

The sharp transition around indices 300--400 (Appendix~\ref{app:linear-approx}), where the singular value spectrum flattens, marks the boundary where the trained model's local linearity begins to dominate. These results reinforce our hypothesis: the directions supporting the longest GIs also align with those where the model behaves most linearly. Training amplifies this structure, enabling extended ray-based coverage without changes in prediction.

\subsection{Effect of Ray Coverage Interventions on Generalization}
\label{sec:rc_validity}

In this section, we show that intervening on ray coverage changes the model's generalization ability, highlighting their close relationship beyond the post-hoc analysis we have so far. In Appendix~\ref{app:causal_derivation}, we show theoretically that, under what we call the \emph{no-re-entry-assumption}, i.e., the assumption that the model does not predict the original class after leaving GI, an increase in GI implies a decrease in the prediction error. This empirical experiment expands on the theoretical result.

The effect of modifying ray coverage is evaluated using an auxiliary loss applied to perturbed training inputs. In the \textbf{expand} condition, correct classification of the perturbed input is encouraged. In the \textbf{restrict} condition, confident misclassification of the perturbed input is encouraged, while cross-entropy is retained at the original training input. The \textbf{control} condition uses cross-entropy alone. An MNIST-MLP and a Fashion-MNIST-CNN are trained under each condition using three random seeds. Full loss definitions and hyperparameters are provided in Appendix~\ref{app:causal_intervention}.

\begin{table}[ht]
\centering
\small
\caption{Test accuracy, mean GI, and ray coverage under control, expansion, and restriction conditions. Values are reported as mean $\pm$ standard deviation over three random seeds at the final training epoch.}
\label{tab:causal_intervention}
\begin{tabular}{llccc}
\toprule
Model & Condition & Test Acc. (\%) & Mean GI & Ray Coverage (\%) \\
\midrule
\multirow{3}{*}{MNIST MLP}
 & Control  & $87.38\pm0.08$ & $1.872\pm0.045$ & $78.33\pm1.04$ \\
 & Expand   & $89.04\pm0.08$ & $1.943\pm0.054$ & $81.17\pm1.26$ \\
 & Restrict & $75.18\pm1.18$ & $1.498\pm0.025$ & $60.67\pm0.76$ \\
\midrule
\multirow{3}{*}{Fashion-MNIST CNN}
 & Control  & $85.27\pm0.18$ & $2.005\pm0.017$ & $73.83\pm0.29$ \\
 & Expand   & $86.37\pm0.29$ & $2.037\pm0.011$ & $74.33\pm0.58$ \\
 & Restrict & $77.86\pm1.29$ & $1.492\pm0.092$ & $54.33\pm3.21$ \\
\bottomrule
\end{tabular}
\end{table}

As shown in Table~\ref{tab:causal_intervention}, coverage restriction reduces test accuracy relative to the control by 12.2 percentage points on MNIST-MLP ($p=0.003$) and 7.4 percentage points on Fashion-MNIST-CNN ($p\approx0.008$). Coverage expansion increases accuracy by 1.7 and 1.1 percentage points, respectively (both $p\leq0.008$). Corresponding changes in GI and ray coverage are observed from the first epoch across all seeds. These intervention results are consistent with the theoretical relationship between GI expansion and population error and support a causal contribution of ray coverage to generalization.

\subsection{Ray Coverage as a Mechanism for Test-Time Classification}

Building on the directional invariance captured by GIs, we show that ray coverage provides a mechanistic explanation for test-time classification decisions. This framework unifies standard generalization, random label learning, long-tailed distribution learning, and adversarial vulnerability under a single geometric perspective.

\subsubsection{Test Sample Classification via Ray Coverage}
We analyze the relationship between test sample classification and the ray coverage formed by training samples across MNIST, CIFAR-10, and ImageNette datasets using multiple architectures. For each test sample, we compute GIs along the direction to all training samples of the predicted class and report the percentage of GIs exceeding 1.0, representing how often the test sample falls within the ray coverage region of the predicted class.

\begin{figure}[t]
\centering
\includegraphics[width=\linewidth]{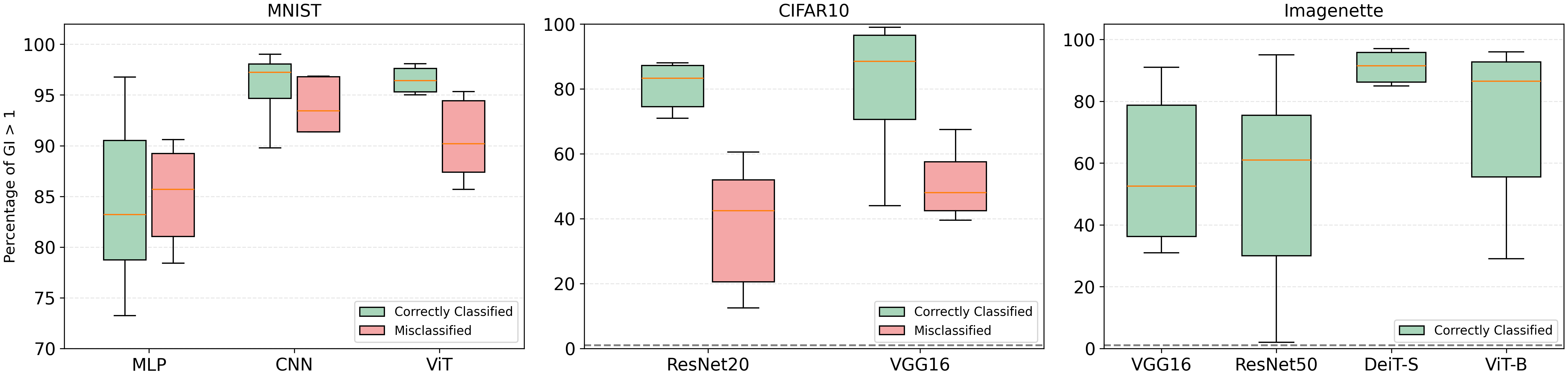}
\caption{Test Sample Classification via Ray Coverage. Box plots show the percentage of GIs exceeding 1.0 between test and training samples of the corresponding class. Both correctly classified and misclassified samples exhibit substantial ray coverage across datasets and architectures. The gray dashed line indicates the threshold for directional coverage, where a test sample is considered covered if at least 1\% of training samples from the same class exhibit a GI greater than 1.0. Misclassified bars are absent for the ImageNette models because no misclassified samples appeared in the evaluated subset (k=20) of test samples per class, likely due to the limited sample size.}
\label{fig:test_samples_ray_coverage}
\end{figure}

We observe consistent patterns across datasets and models; Figure~\ref{fig:test_samples_ray_coverage} reveals consistently high coverage across datasets and architectures for both correctly classified and misclassified samples.

These findings suggest that test-time predictions arise from the geometry of class-level ray coverage rather than instance-level memorization. Even misclassified inputs often fall within the invariant region of the predicted class, supporting Neural Expansion as a general mechanism underlying classification across varying accuracy regimes. We include more details about the experiment in Appendix~\ref{app:test-to-train-protocol}. Class trajectory visualizations along train-to-test paths are shown in Appendix~\ref{app:class_region_analysis}. These measurements exhibit seed stability and class-discriminability (Appendices~\ref{app:rc_discriminative}, \ref{app:rc_stat_robustness}).

\subsubsection{Neural Expansion as a Unified Framework for Challenging Cases}
We examine whether ray coverage remains the operative mechanism in challenging settings of traditional learning: random label learning, long-tailed distributions, and adversarial perturbations.
\begin{figure}[t]
\centering
\includegraphics[width=\linewidth]{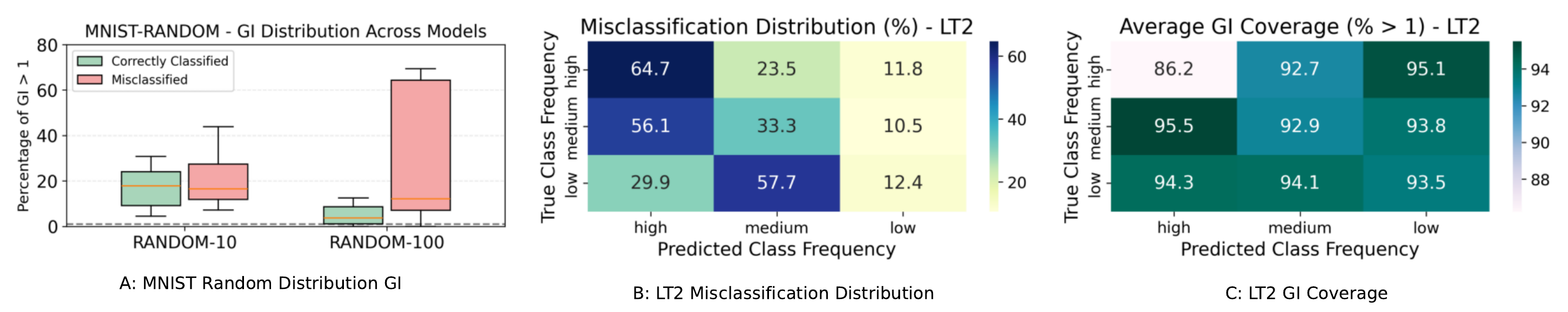}
\caption{\textbf{Left:} Random Label Learning with GI ray coverage for RANDOM-10 and RANDOM-100 models. The gray dashed line (as defined in Figure~\ref{fig:test_samples_ray_coverage}) indicates the 1\% directional coverage threshold. \textbf{Middle:} LT2 Misclassification Distribution showing percentage distribution from true class frequency to predicted class frequency. \textbf{Right:} LT2 GI Coverage showing average percentage of GIs exceeding 1.0 across different frequency categories.}
\label{fig:random_longtail}
\end{figure}

\begin{figure}[t]
\centering
\includegraphics[width=\linewidth]{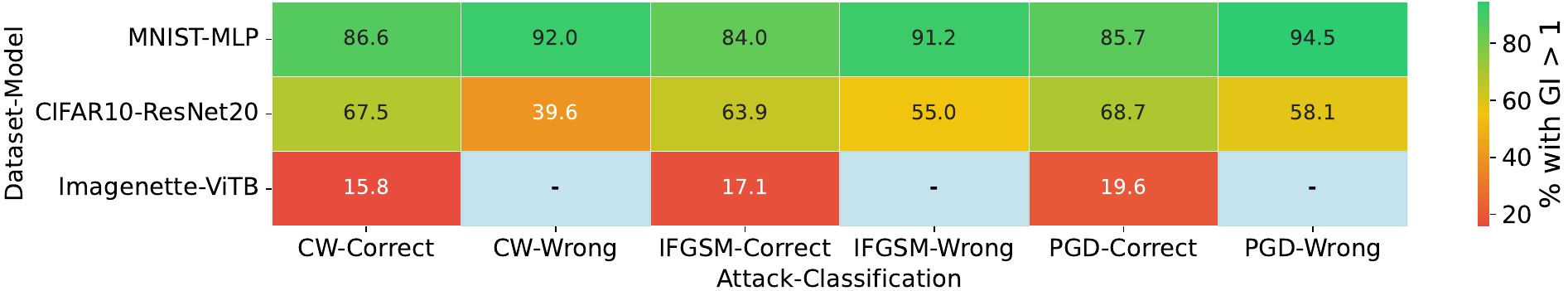}
\caption{Adversarial Examples Analysis (misclassified samples indicate successful attacks; correct classifications indicate failures). The heatmap shows the percentage of adversarial examples with GI > 1 to training samples of the predicted class across different dataset-model combinations and attack methods (CW, IFGSM, PGD).}
\label{fig:adversarial}
\end{figure}

\textbf{Random Label Learning:} Even when label semantics are randomized (RANDOM-10, RANDOM-100~\footnote{Since the labels are randomized, we are free to choose the total number of classes as well.}), we find that test samples consistently fall within ray coverage zones around training samples with the same assigned label. The plot in Figure~\ref{fig:random_longtail}a reveals nontrivial GI overlap between test and training samples of the same (randomly assigned) label, indicating that the model still forms geometric zones for each label, albeit smaller and noisier. This finding challenges the common assumption that networks simply ``memorize'' random labels without forming structure. (See Appendix~\ref{app:random_label} for dataset-specific coverage values.)

\textbf{Long-Tailed Distributions:} In highly imbalanced datasets (e.g., MNIST-LT2), test samples from underrepresented classes are disproportionately misclassified into medium- or high-frequency classes (Figure~\ref{fig:random_longtail}b). The ray coverage plot in Figure~\ref{fig:random_longtail}c shows that this is not due to failure in learning low-frequency classes, but rather to the expansive coverage of frequent classes dominating the input space. Misclassifications align with the regions where dominant classes extend their influence. (Detailed statistics in Appendix~\ref{app:long_tail_distribution}.)

\textbf{Adversarial Examples:} We analyze how adversarial examples interact with ray coverage using three attack methods: Carlini-Wagner (CW)~\citep{carlini2017evaluatingrobustnessneuralnetworks}, Iterative Fast Gradient Sign Method (IFGSM)~\citep{kurakin2017adversarialexamplesphysicalworld}, and Projected Gradient Descent (PGD)~\citep{madry2019deeplearningmodelsresistant}. Figure~\ref{fig:adversarial} reveals the consistent presence of ray coverage regardless of whether adversarial perturbations successfully change the classification. In many cases, the attack appears to shift the test input from one class's coverage zone to another's, rather than entirely escaping the structure. This trend persists across MNIST and CIFAR-10.

For ImageNette-ViTB, correctly classified adversarial examples show low ray coverage. Notably, we could not compute coverage values for misclassified ImageNette examples, as they were often pushed outside our 10-class subset into the broader 1000-class ImageNet space. This illustrates how high-dimensional natural images provide more pathways for adversarial examples to escape class boundaries than constrained datasets like MNIST. Appendix~\ref{app:adversarial_examples_experiments} provides more details regarding the geometry of adversarial perturbations. To further examine model behavior under distribution shift, we also compute GI coverage for out-of-distribution samples (see Appendix~\ref{app:ood_sample_analysis}). The anisotropic growth of GIs during training, driven by margin increase and directional restructuring of Jacobian singular values, is analyzed in Appendix~\ref{sec:gi_growth_understanding}.

\section{Discussion}
Our results reveal a striking finding that cuts across all experimental conditions: ray coverage operates as the classification mechanism even when it \emph{should not}. Networks trained on random labels still form geometric coverage zones, and misclassified samples still fall within ray coverage, just of the wrong class. This suggests that gradient descent does not merely fit labels; it \emph{consistently} organizes input space into directional corridors, regardless of whether those corridors reflect semantic structure, though these corridors become significantly sparser and more fragmented without semantic supervision (17\%--24\% coverage under random labels versus $>$80\% on standard datasets, Table~\ref{tab:random_labels}). This clarifies the boundary between memorization and generalization: both operate through ray coverage, but generalizing networks build wider, overlapping corridors aligned with low-curvature directions that capture class-relevant structure.

This mechanistic picture complements existing theories in ways that are non-redundant. PAC-Bayesian~\citep{neyshabur2017pac} and margin-based bounds~\citep{bartlett2017spectrally} certify that a model generalizes globally, but cannot say \emph{why} a specific test point is classified correctly. NTK theory~\citep{jacot2018neural} and Neural Collapse~\citep{papyan2020prevalence} describe either the initialization regime or the terminal phase; our framework tracks the geometry continuously throughout training, filling a sample-level explanatory gap that global theories leave open.

\textbf{Limitations and Future Work.} GI computation remains expensive for high-dimensional inputs, and while our sampling-based approach scales to architectures such as ViTs, full coverage requires further approximation strategies. The framework is currently most natural for classification; extending it to regression, generation, or structured prediction requires reformulating GIs for settings without discrete decision boundaries. A promising future direction is to integrate these insights directly into training objectives, encouraging networks to explicitly maximize ray coverage in low-curvature directions to enhance generalization and robustness. Finally, the consistent emergence of ray coverage across model classes suggests that gradient-based learning may implicitly align with the intrinsic geometry of natural data distributions, a connection that could yield new principles for representation learning and regularization. A rigorous proof that individual Jacobian singular values restructure anisotropically during gradient descent training remains an open theoretical question; our current account in Section~\ref{sec:gi_growth_understanding} relies on empirical observations.

\section{Conclusion}
We present Neural Expansion, a geometric framework that reinterprets generalization in deep neural networks through the lens of directionally-sensitive invariance pathways. Through extensive empirical analysis spanning diverse architectures, datasets, and training regimes, we uncover a coherent geometric signature: a compact set of high-curvature directions sculpts class boundaries, while an abundance of low-curvature directions carve out invariance corridors that link samples belonging to the same class. This dual structure provides a unifying account of seemingly disparate phenomena, namely standard generalization, memorization of random labels, misclassification of tail-class samples, and susceptibility to adversarial perturbations, all through a common geometric lens. The resulting picture reframes our understanding of what neural networks learn: rather than constructing intricate decision surfaces, they organically develop stable classification pathways whose collective geometry mirrors the latent structure of the data.

\newpage

\bibliographystyle{apalike}
\bibliography{refs}

@article{bartlett2017spectrally,
  title={Spectrally-normalized margin bounds for neural networks},
  author={Bartlett, Peter L and Foster, Dylan J and Telgarsky, Matus J},
  journal={Advances in neural information processing systems},
  volume={30},
  year={2017}
}

@inproceedings{keskar2017largebatchtrainingdeeplearning,
  title={On Large-Batch Training for Deep Learning: Generalization Gap and Sharp Minima},
  author={Nitish Shirish Keskar and Dheevatsa Mudigere and Jorge Nocedal and Mikhail Smelyanskiy and Ping Tak Peter Tang},
  booktitle={International Conference on Learning Representations},
  year={2017}
}

@article{neyshabur2017pac,
  title={A pac-bayesian approach to spectrally-normalized margin bounds for neural networks},
  author={Neyshabur, Behnam and Bhojanapalli, Srinadh and Srebro, Nathan},
  journal={arXiv preprint arXiv:1707.09564},
  year={2017}
}

@inproceedings{arora2018stronger,
  title={Stronger generalization bounds for deep nets via a compression approach},
  author={Arora, Sanjeev and Ge, Rong and Neyshabur, Behnam and Zhang, Yi},
  booktitle={International conference on machine learning},
  pages={254--263},
  year={2018},
  organization={PMLR}
}

@article{jacot2018neural,
  title={Neural tangent kernel: Convergence and generalization in neural networks},
  author={Jacot, Arthur and Gabriel, Franck and Hongler, Cl{\'e}ment},
  journal={Advances in neural information processing systems},
  volume={31},
  year={2018}
}

@article{chizat2019lazy,
  title={On lazy training in differentiable programming},
  author={Chizat, Lenaic and Oyallon, Edouard and Bach, Francis},
  journal={Advances in neural information processing systems},
  volume={32},
  year={2019}
}

@article{papyan2020prevalence,
  title={Prevalence of neural collapse during the terminal phase of deep learning training},
  author={Papyan, Vardan and Han, XY and Donoho, David L},
  journal={Proceedings of the National Academy of Sciences},
  volume={117},
  number={40},
  pages={24652--24663},
  year={2020},
  publisher={National Academy of Sciences}
}

@article{han2021neural,
  title={Neural collapse under mse loss: Proximity to and dynamics on the central path},
  author={Han, XY and Papyan, Vardan and Donoho, David L},
  journal={arXiv preprint arXiv:2106.02073},
  year={2021}
}

@article{wu2024linguistic,
  title={Linguistic collapse: Neural collapse in (large) language models},
  author={Wu, Robert and Papyan, Vardan},
  journal={Advances in Neural Information Processing Systems},
  volume={37},
  pages={137432--137473},
  year={2024}
}

@inproceedings{valle2018deep,
  title={Deep Learning Generalizes Because the Parameter-Function Map Is Biased Towards Simple Functions},
  author={Valle-Perez, Guillermo and Camargo, Chico Q and Louis, Ard A},
  booktitle={International Conference on Learning Representations},
  year={2019}
}

@article{belkin1812reconciling,
  title={Reconciling modern machine learning practice and the bias-variance trade-off},
  author={Belkin, Mikhail and Hsu, Daniel and Ma, Siyuan and Mandal, Soumik},
  journal={arXiv preprint arXiv:1812.11118 v2},
  volume={14},
  year={2018}
}

@inproceedings{zhang2016understanding,
  title={Understanding Deep Learning Requires Rethinking Generalization},
  author={Zhang, Chiyuan and Bengio, Samy and Hardt, Moritz and Recht, Benjamin and Vinyals, Oriol},
  booktitle={International Conference on Learning Representations},
  year={2017}
}

@article{novak2018sensitivity,
  title={Sensitivity and generalization in neural networks: an empirical study},
  author={Novak, Roman and Bahri, Yasaman and Abolafia, Daniel A and Pennington, Jeffrey and Sohl-Dickstein, Jascha},
  journal={arXiv preprint arXiv:1802.08760},
  year={2018}
}

@article{beaglehole2024average,
  title={Average gradient outer product as a mechanism for deep neural collapse},
  author={Beaglehole, Daniel and S{\'u}ken{\'\i}k, Peter and Mondelli, Marco and Belkin, Misha},
  journal={Advances in Neural Information Processing Systems},
  volume={37},
  pages={130764--130796},
  year={2024}
}

@inproceedings{adlam2020neural,
  title={The neural tangent kernel in high dimensions: Triple descent and a multi-scale theory of generalization},
  author={Adlam, Ben and Pennington, Jeffrey},
  booktitle={International Conference on Machine Learning},
  pages={74--84},
  year={2020},
  organization={PMLR}
}

@article{nakkiran2021deep,
  title={Deep double descent: Where bigger models and more data hurt},
  author={Nakkiran, Preetum and Kaplun, Gal and Bansal, Yamini and Yang, Tristan and Barak, Boaz and Sutskever, Ilya},
  journal={Journal of Statistical Mechanics: Theory and Experiment},
  volume={2021},
  number={12},
  pages={124003},
  year={2021},
  publisher={IOP Publishing}
}

@inproceedings{woodworth2020kernel,
  title={Kernel and rich regimes in overparametrized models},
  author={Woodworth, Blake and Gunasekar, Suriya and Lee, Jason D and Moroshko, Edward and Savarese, Pedro and Golan, Itay and Soudry, Daniel and Srebro, Nathan},
  booktitle={Conference on Learning Theory},
  pages={3635--3673},
  year={2020},
  organization={PMLR}
}

@inproceedings{deng2009imagenet,
  title={Imagenet: A large-scale hierarchical image database},
  author={Deng, Jia and Dong, Wei and Socher, Richard and Li, Li-Jia and Li, Kai and Fei-Fei, Li},
  booktitle={2009 IEEE conference on computer vision and pattern recognition},
  pages={248--255},
  year={2009},
  organization={Ieee}
}

@techreport{krizhevsky2009learning,
  title={Learning Multiple Layers of Features from Tiny Images},
  author={Krizhevsky, Alex and Hinton, Geoffrey},
  institution={University of Toronto},
  year={2009}
}

@article{lecun1998gradient,
  title={Gradient-based learning applied to document recognition},
  author={LeCun, Yann and Bottou, L{\'e}on and Bengio, Yoshua and Haffner, Patrick},
  journal={Proceedings of the IEEE},
  volume={86},
  number={11},
  pages={2278--2324},
  year={1998},
  publisher={IEEE}
}

@article{ma2022delving,
  title={Delving into semantic scale imbalance},
  author={Ma, Yanbiao and Jiao, Licheng and Liu, Fang and Li, Yuxin and Yang, Shuyuan and Liu, Xu},
  journal={arXiv preprint arXiv:2212.14613},
  year={2022}
}

@inproceedings{carlini2017evaluatingrobustnessneuralnetworks,
  title={Towards Evaluating the Robustness of Neural Networks},
  author={Nicholas Carlini and David Wagner},
  booktitle={2017 IEEE Symposium on Security and Privacy (SP)},
  pages={39--57},
  year={2017},
  organization={IEEE}
}

@inproceedings{kurakin2017adversarialexamplesphysicalworld,
  title={Adversarial Examples in the Physical World},
  author={Alexey Kurakin and Ian Goodfellow and Samy Bengio},
  booktitle={ICLR Workshop},
  year={2017}
}

@inproceedings{madry2019deeplearningmodelsresistant,
  title={Towards Deep Learning Models Resistant to Adversarial Attacks},
  author={Aleksander Madry and Aleksandar Makelov and Ludwig Schmidt and Dimitris Tsipras and Adrian Vladu},
  booktitle={International Conference on Learning Representations},
  year={2018}
}

@InProceedings{He_2016_CVPR,
author = {He, Kaiming and Zhang, Xiangyu and Ren, Shaoqing and Sun, Jian},
title = {Deep Residual Learning for Image Recognition},
booktitle = {Proceedings of the IEEE Conference on Computer Vision and Pattern Recognition (CVPR)},
month = {June},
year = {2016}
}

@inproceedings{simonyan2015deepconvolutionalnetworkslargescale,
  title={Very Deep Convolutional Networks for Large-Scale Image Recognition},
  author={Karen Simonyan and Andrew Zisserman},
  booktitle={International Conference on Learning Representations},
  year={2015}
}

@inproceedings{salman2020towards,
  title={Towards Quantifying Intrinsic Generalization of Deep ReLU Networks},
  author={Salman, Shaeke and Zhang, Canlin and Liu, Xiuwen and Mio, Washington},
  booktitle={2020 International Joint Conference on Neural Networks (IJCNN)},
  pages={1--9},
  year={2020},
  organization={IEEE}
}

@article{bengio2013representation,
  title={Representation learning: A review and new perspectives},
  author={Bengio, Yoshua and Courville, Aaron and Vincent, Pascal},
  journal={IEEE transactions on pattern analysis and machine intelligence},
  volume={35},
  number={8},
  pages={1798--1828},
  year={2013},
  publisher={IEEE}
}

@inproceedings{dosovitskiy2020image,
  title={An Image is Worth 16x16 Words: Transformers for Image Recognition at Scale},
  author={Dosovitskiy, Alexey and Beyer, Lucas and Kolesnikov, Alexander and Weissenborn, Dirk and Zhai, Xiaohua and Unterthiner, Thomas and Dehghani, Mostafa and Minderer, Matthias and Heigold, Georg and Gelly, Sylvain and others},
  booktitle={International Conference on Learning Representations},
  year={2021}
}

@inproceedings{
Lyu2020Gradient,
title={Gradient Descent Maximizes the Margin of Homogeneous Neural Networks},
author={Kaifeng Lyu and Jian Li},
booktitle={International Conference on Learning Representations},
year={2020},
url={https://openreview.net/forum?id=SJeLIgBKPS}
}

@inproceedings{foret2020sharpness,
  title={Sharpness-Aware Minimization for Efficiently Improving Generalization},
  author={Foret, Pierre and Kleiner, Ariel and Mobahi, Hossein and Neyshabur, Behnam},
  booktitle={International Conference on Learning Representations},
  year={2021}
}

@article{vardi2023implicit,
  title={On the implicit bias in deep-learning algorithms},
  author={Vardi, Gal},
  journal={Communications of the ACM},
  volume={66},
  number={6},
  pages={86--93},
  year={2023},
  publisher={ACM New York, NY, USA}
}

@inproceedings{frei2023benign,
  title={Benign overfitting in linear classifiers and leaky relu networks from kkt conditions for margin maximization},
  author={Frei, Spencer and Vardi, Gal and Bartlett, Peter and Srebro, Nathan},
  booktitle={The Thirty Sixth Annual Conference on Learning Theory},
  pages={3173--3228},
  year={2023},
  organization={PMLR}
}

@inproceedings{andriushchenko2022sam,
  title={Towards Understanding Sharpness-Aware Minimization},
  author={Andriushchenko, Maksym and Flammarion, Nicolas},
  booktitle={Proceedings of the 39th International Conference on Machine Learning},
  pages={639--668},
  year={2022},
  organization={PMLR}
}

@inproceedings{andriushchenko2023lowrank,
  title={Sharpness-Aware Minimization Leads to Low-Rank Features},
  author={Andriushchenko, Maksym and Bahri, Dara and Mobahi, Hossein and Flammarion, Nicolas},
  booktitle={Advances in Neural Information Processing Systems},
  volume={36},
  year={2023}
}

@inproceedings{movahedi2025geometric,
  title={Geometric Inductive Biases of Deep Networks: The Role of Data and Architecture},
  author={Movahedi, Sajad and others},
  booktitle={International Conference on Learning Representations},
  year={2025},
  note={Spotlight}
}

\clearpage
\appendix

\section{Detailed Experimental Setup}
\label{app:experimental_details}

\subsection{Datasets and Variants}
\label{app:datasets_and_variants}

We use three benchmark datasets: \textbf{MNIST}~\citep{lecun1998gradient} ($28 \times 28$ grayscale, 10 classes), \textbf{CIFAR-10}~\citep{krizhevsky2009learning} ($32 \times 32$ RGB, 10 classes), and \textbf{ImageNette}~\citep{deng2009imagenet}, a 10-class subset of ImageNet with $224 \times 224$ RGB images.

To test generalization under different distributions, we created variants including:
\begin{itemize}
    \item \textbf{Random Label Variants}: RANDOM-10 (random labels across 10 classes) and RANDOM-100 (random labels across 100 classes), following~\citet{zhang2016understanding}.
    \item \textbf{Long-tailed Variants}: MNIST-LT1 (imbalance ratio 6.5:1) and MNIST-LT2 (78.6:1), following~\citet{ma2022delving}.
\end{itemize}

Images are normalized to $[-1, 1]$ using dataset-specific preprocessing. For ImageNette, we apply standard ImageNet normalization, random cropping, horizontal flipping, and color jitter.

Their respective open licenses: MIT (CIFAR-10), CC BY-SA 3.0 (MNIST), and Apache 2.0 (Imagenette, ViT).

\subsection{Model Architectures}
\label{app:model_architectures}

We evaluate three architecture families:

\begin{itemize}
    \item \textbf{Multi-Layer Perceptrons (MLPs)}: For MNIST, we use a 784--1024--512--10 MLP with ReLU and batch normalization. For CIFAR-10, a 3072--2048--1024--10 MLP is used.

    \item \textbf{Convolutional Neural Networks (CNNs)}: We use custom CNNs for MNIST and CIFAR-10, along with pre-trained ResNet-20, ResNet-50, and VGG-16~\citep{He_2016_CVPR,simonyan2015deepconvolutionalnetworkslargescale}.

    \item \textbf{Vision Transformers (ViTs)}: Lightweight ViTs for MNIST (patch size $7 \times 7$, embedding dim 64) and CIFAR-10 (patch size $8 \times 8$, embedding dim 192). For ImageNette, we use ViT-B/16 and DeiT-S~\citep{dosovitskiy2020image}.
\end{itemize}

\subsection{Training Protocols}

All custom models are trained using SGD until convergence with dataset-specific hyperparameters:

\begin{itemize}
    \item \textbf{MNIST}: 30 epochs, learning rate $10^{-3}$/$10^{-2}$, no momentum or weight decay.
    \item \textbf{CIFAR-10}: 50 epochs, learning rate $10^{-3}$/$10^{-2}$, momentum 0.9, weight decay $10^{-4}$.
    \item \textbf{ImageNette}: 30 epochs, learning rate $10^{-4}$, momentum 0.9, weight decay $10^{-4}$.
\end{itemize}

\begin{figure}[ht]
    \centering
    \begin{subfigure}[t]{0.32\textwidth}
        \includegraphics[width=\linewidth]{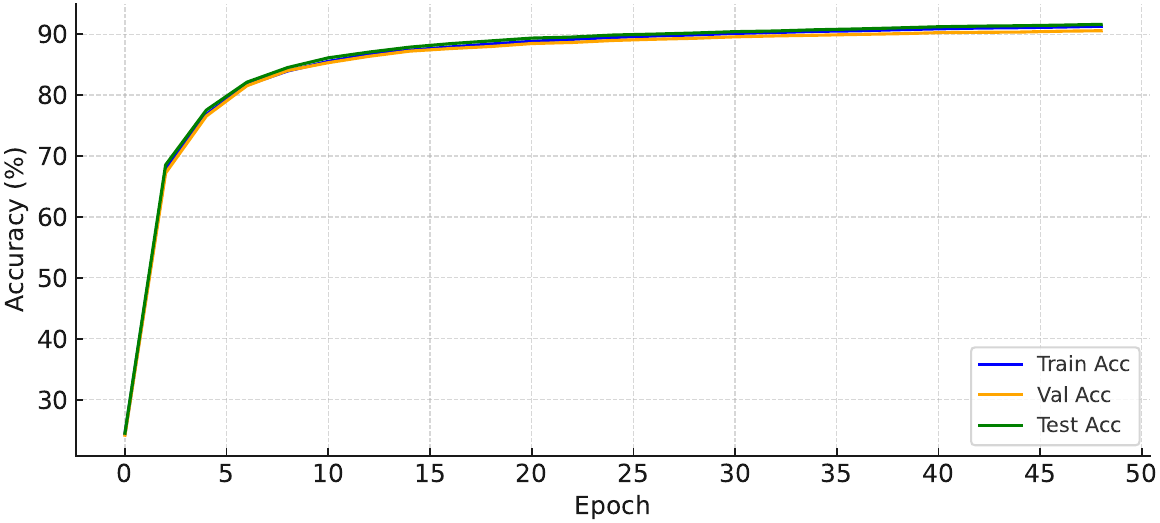}
        \caption{MNIST MLP}
    \end{subfigure}
    \begin{subfigure}[t]{0.32\textwidth}
        \includegraphics[width=\linewidth]{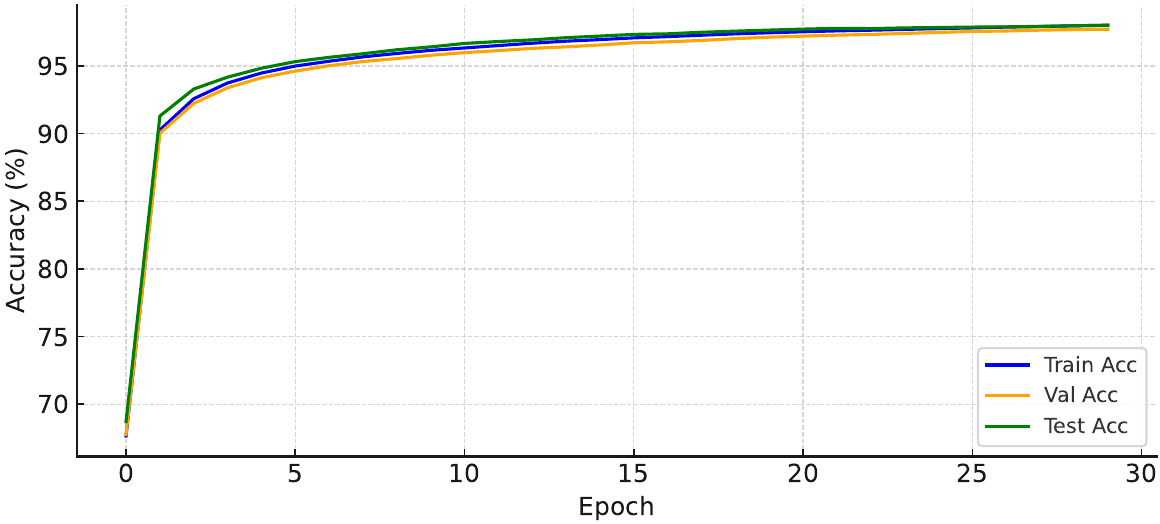}
        \caption{MNIST CNN}
    \end{subfigure}
    \begin{subfigure}[t]{0.32\textwidth}
        \includegraphics[width=\linewidth]{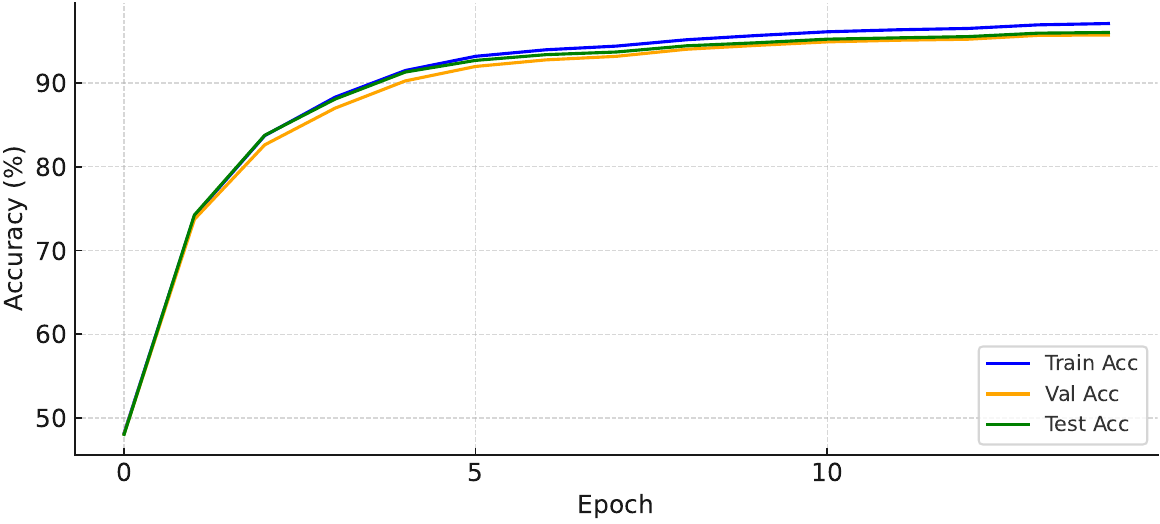}
        \caption{MNIST ViT}
    \end{subfigure}
    \vspace{0.5em}
    \begin{subfigure}[t]{0.32\textwidth}
        \includegraphics[width=\linewidth]{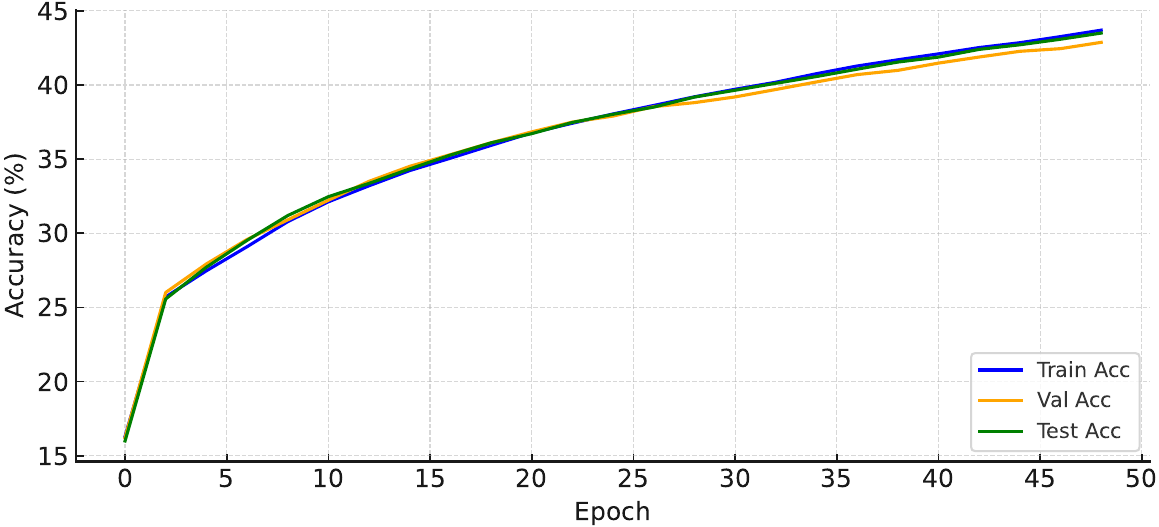}
        \caption{CIFAR-10 MLP}
    \end{subfigure}
    \begin{subfigure}[t]{0.32\textwidth}
        \includegraphics[width=\linewidth]{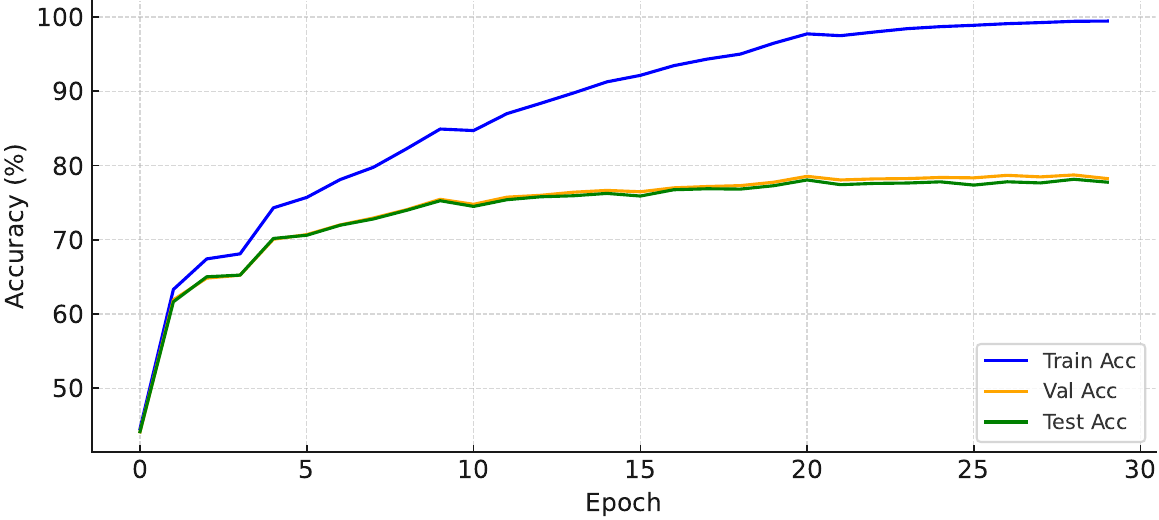}
        \caption{CIFAR-10 CNN}
    \end{subfigure}
    \begin{subfigure}[t]{0.32\textwidth}
        \includegraphics[width=\linewidth]{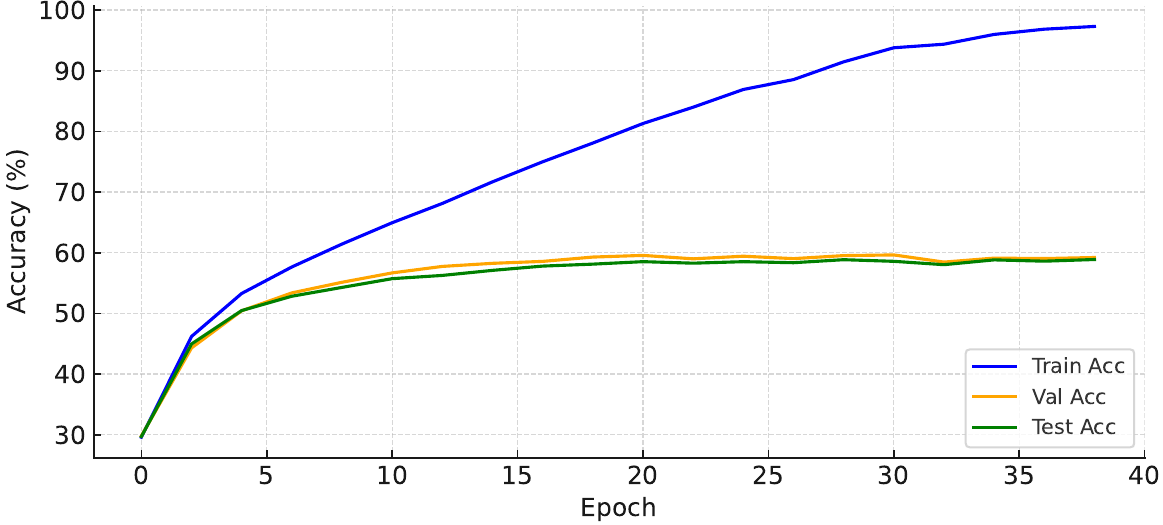}
        \caption{CIFAR-10 ViT}
    \end{subfigure}
    \vspace{0.5em}
    \begin{subfigure}[t]{0.24\textwidth}
        \includegraphics[width=\linewidth]{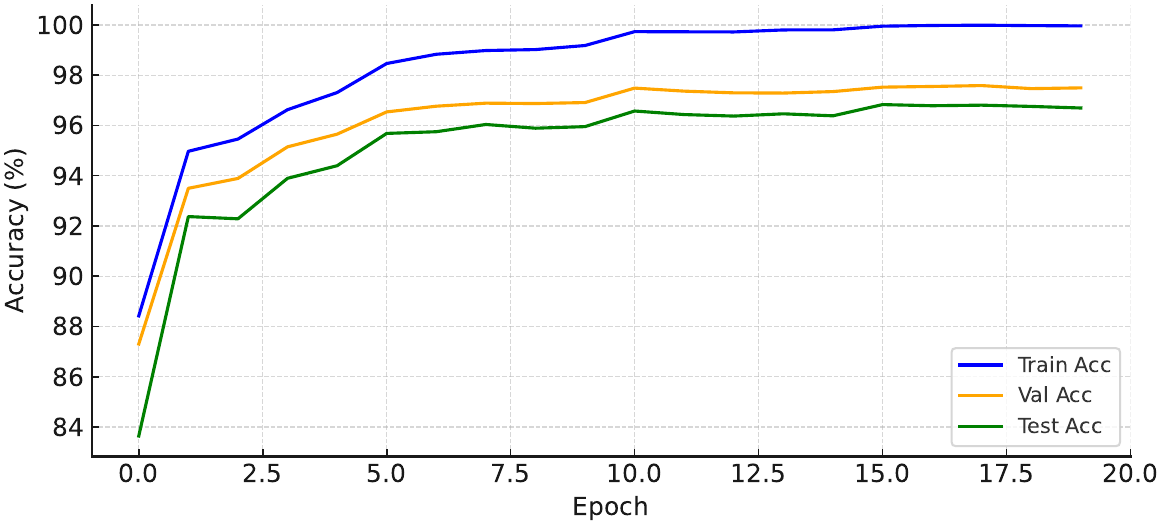}
        \caption{MNIST MLP LT1}
    \end{subfigure}
    \begin{subfigure}[t]{0.24\textwidth}
        \includegraphics[width=\linewidth]{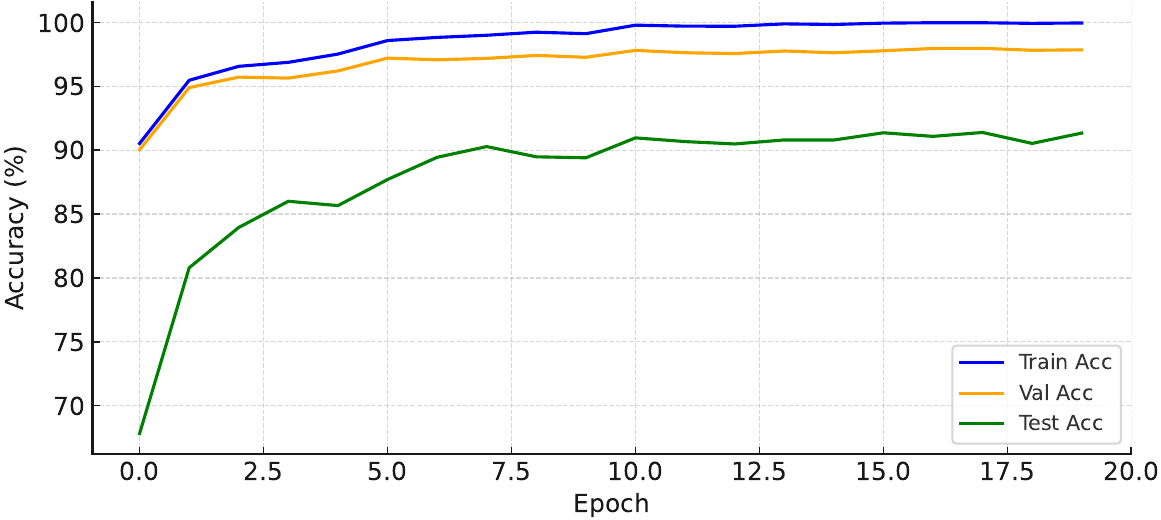}
        \caption{MNIST MLP LT2}
    \end{subfigure}
    \begin{subfigure}[t]{0.24\textwidth}
        \includegraphics[width=\linewidth]{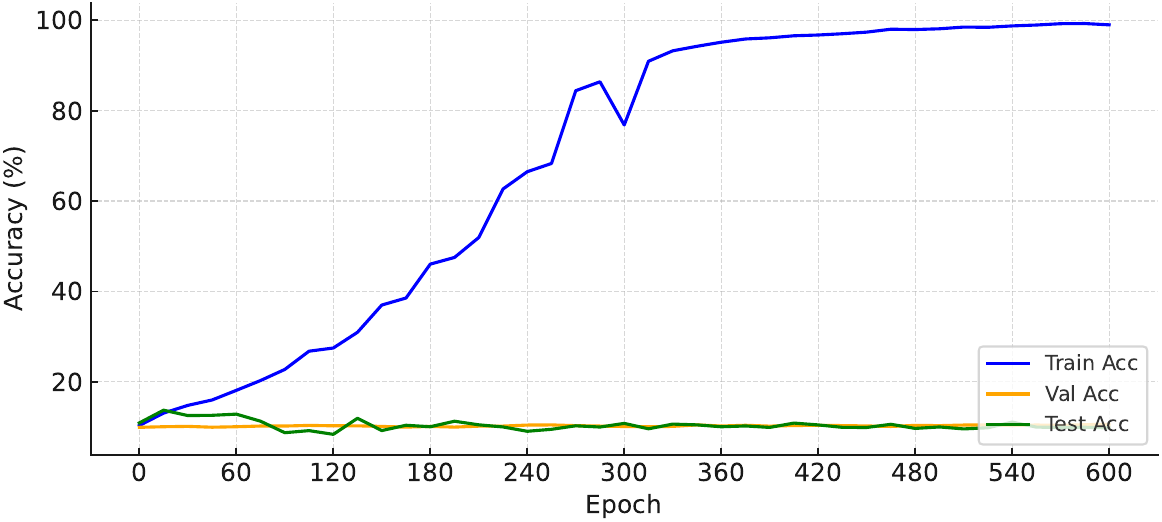}
        \caption{MLP Random-10}
    \end{subfigure}
    \begin{subfigure}[t]{0.24\textwidth}
        \includegraphics[width=\linewidth]{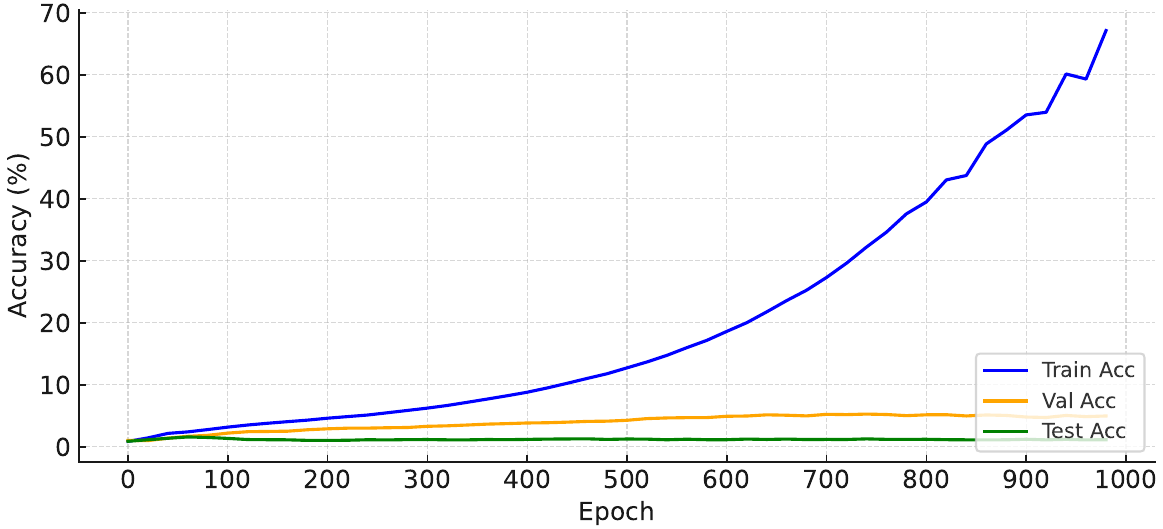}
        \caption{MLP Random-100}
    \end{subfigure}
    \caption{Epoch-wise accuracy curves across models and datasets for vanilla as well as LT1, LT2, Random-10 and Random-100 variations.}
    \label{fig:accuracy-original}
\end{figure}

We include accuracy for all the custom-trained models in Figure~\ref{fig:accuracy-original}.

\subsection{Analysis Methods}
\label{app:analysis-methods}
\textbf{Generalization Interval Calculation}: We implemented a step-wise search algorithm to compute GIs between sample pairs. For direction vector $\mathbf{v} = \mathbf{x}_2 - \mathbf{x}_1$ (the unnormalized difference vector), we incrementally moved along both positive and negative directions ($\mathbf{x}_1 \pm \epsilon\mathbf{v}$) using step size $\delta=0.001$-$0.01$ until either a decision boundary was crossed or a maximum $\epsilon=3.0$ was reached. Under this convention, $\epsilon = 1.0$ exactly reaches $\mathbf{x}_2$, and the threshold $\mathrm{GI}(f, x_i, x_{\text{test}} - x_i) \geq 1$ in Equation~\eqref{eq:coverage} is dimensionally consistent with the unnormalized direction. Our vectorized implementation enabled batch processing (MNIST: 6000 pairs, CIFAR-10: 2000 pairs, ImageNette: 100 pairs). Table~\ref{tab:gi_compute_summary} contains the compute resources and runtime details of the GI coverage experiments.

\begin{table}[ht]
\centering
\caption{Compute resources and runtime details for test-to-train GI coverage experiments using a step-wise directional search.}
\label{tab:gi_compute_summary}
\resizebox{\textwidth}{!}{\begin{tabular}{lcccc}
\toprule
\textbf{Dataset} & \textbf{Architecture} & \textbf{GPU(s)} & \textbf{Runtime (hh:mm)} & \textbf{Memory Used (GB)} \\
\midrule
\multirow{3}{*}{MNIST (6000 pairs)}
  & \textbf{MLP}                       & A5000 & 00:30 & 3.0 \\
  & \textbf{CNN}                       & A5000 & 00:55 & 5.0 \\
  & \textbf{ViT}                       & A5000 & 01:25 & 6.5 \\
\midrule
\multirow{2}{*}{CIFAR-10 (2000 pairs)}
  & \textbf{ResNet20}                  & A5000 & 00:40 & 4.2 \\
  & \textbf{VGG16-BN}                  & A5000 & 00:55 & 5.1 \\
\midrule
\multirow{4}{*}{ImageNette (100 pairs)}
  & \textbf{VGG16}                    & A5000 & 01:40 & 6.8 \\
  & \textbf{ResNet50}                 & A5000 & 01:30 & 6.5 \\
  & \textbf{ViT-B/16}                 & A5000 & 01:55 & 7.8 \\
  & \textbf{DeiT-Small}               & A5000 & 01:35 & 7.0 \\
\bottomrule
\end{tabular}}
\end{table}

\textbf{Jacobian Analysis}: For each sample, we computed the Jacobian matrix $J = \frac{\partial f(x)}{\partial x}$ of the penultimate layer features with respect to the input and performed SVD: $J = U\Sigma V^T$. We evaluated linear approximation quality by measuring cosine similarity between $f(x_0 + \alpha v_i)$ and $f(x_0) + \alpha W\sigma_i u_i$ for different $\alpha$ values, where both terms are $c$-dimensional logit vectors.

\textbf{Ray Coverage Analysis}: For each test sample with predicted class $c$, we calculated GIs to training samples of class $c$ and computed the percentage exceeding 1.0. This quantified test samples' membership in training samples' ray coverage zones, enabling comparisons between correct and incorrect classifications across datasets, architectures, and distribution variants.

\section{GI Evolution Experimental Details}
\label{app:gi_evolution}

To analyze the temporal evolution of ray coverage during training, we tracked GI development in an MNIST MLP (784-1024-512-10) across 10 epochs. Figure~\ref{fig:gi_evolution} captures three key aspects of this evolution.

For train-to-train GI analysis (left panel), we calculated GIs between 100 randomly selected same-class training sample pairs for each digit class at checkpoints from epoch 0 to epoch 10. We measured both mean GI values and the percentage exceeding 1.0, which grew from 0.76 to 1.24 and 39\% to 82\% respectively, exhibiting strong correlation with test accuracy improvement (86.2\% to 91.7\%).

For directional analysis (middle and right panels), we computed GIs along RSVs of the Jacobian matrix and along random directions, using the same scaling factor of 56.0 for both to ensure fair comparison. This factor was derived from the maximum possible distance in the 784-dimensional normalized input space of MNIST. Along RSV directions, mean GI increased from 0.53 to 1.18, with the percentage exceeding 1.0 growing from 22\% to 78\%. In contrast, random directions showed more modest growth from 0.40 to 0.82, with the percentage exceeding 1.0 reaching only 34\% by epoch 10.

\begin{figure}[ht]
\centering
\includegraphics[width=0.8\linewidth]{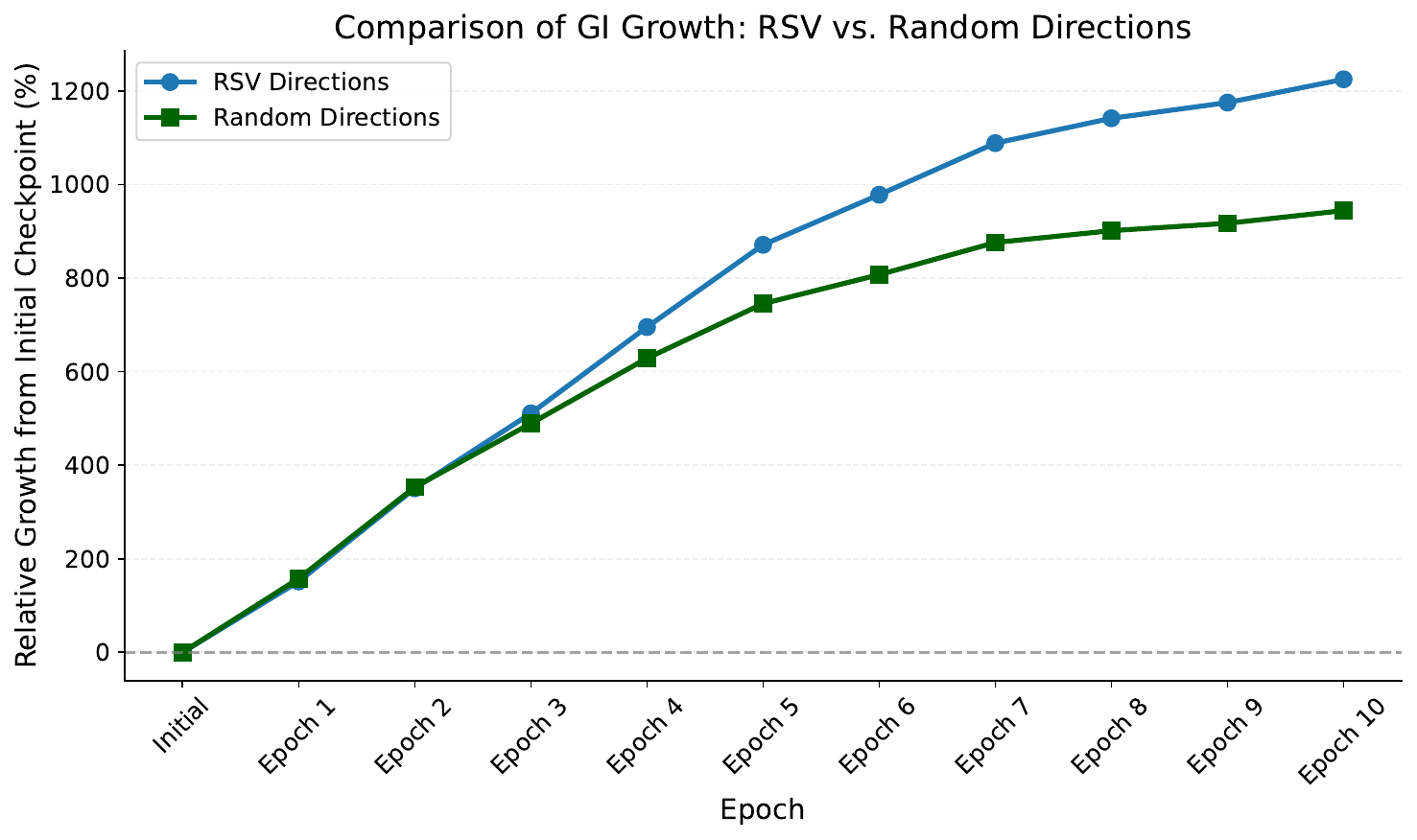}
\caption{Relative GI growth along RSV vs. random directions during training. Both direction types show similar growth patterns through epoch 3, after which RSV directions exhibit substantially accelerated expansion, reaching 1220\% growth by epoch 10 compared to 940\% for random directions.}
\label{fig:gi_growth_comparison}
\end{figure}

Figure~\ref{fig:gi_growth_comparison} provides further insight into the differential growth rates of GIs along RSV versus random directions. When normalized to show relative growth from the initial checkpoint, both direction types exhibit similar expansion through epoch 3 (approximately 500\% growth). However, from epoch 4 onward, a clear divergence emerges, with RSV directions showing accelerated expansion. By epoch 10, RSV directions achieve 1220\% growth relative to initialization, while random directions reach only 940\%. This growing disparity indicates the network progressively develops anisotropic ray coverage that preferentially expands along directions aligned with the underlying data manifold.

To complement these directional and train-to-train analyses, we further examined how ray coverage evolves \emph{within the training set}. At each checkpoint, we computed GIs between randomly selected same-class training sample pairs. Table~\ref{tab:gi_checkpoint_summary} summarizes GI statistics separately for training pairs where the prediction remains correct versus incorrect.

\begin{table}[ht]
\centering
\caption{Summary of GI statistics across training checkpoints for correctly and incorrectly classified training sample pairs. \textit{Mean GI} is the arithmetic mean of all GI values including those at the boundary ($\mathrm{GI}=0$); \textit{Avg GI} is the mean computed only over sample pairs with $\mathrm{GI} > 0$ (i.e., where a nonzero interval exists).}
\label{tab:gi_checkpoint_summary}
\resizebox{\textwidth}{!}{\begin{tabular}{cccccccccccc}
\toprule
Epoch & Batch & \multicolumn{5}{c}{Correct} & \multicolumn{5}{c}{Incorrect} \\
\cmidrule(lr){3-7} \cmidrule(lr){8-12}
 & & Count & Mean GI & \% GI > 1 & Avg GI & Median GI & Count & Mean GI & \% GI > 1 & Avg GI & Median GI \\
\midrule
0  & 0   & 50 & 1.1798 & 48.21 & 1.1638 & 0.41 & 50 & 1.1922 & 48.49 & 1.1816 & 0.49 \\
0  & 200 & 50 & 1.4496 & 79.60 & 1.4708 & 1.38 & 50 & 1.5397 & 84.00 & 1.5635 & 1.48 \\
0  & 399 & 50 & 1.3748 & 80.06 & 1.3850 & 1.21 & 50 & 1.3375 & 82.87 & 1.3553 & 1.28 \\
1  & 399 & 50 & 1.4358 & 74.89 & 1.4523 & 1.23 & 50 & 1.2922 & 81.24 & 1.3065 & 1.30 \\
3  & 200 & 50 & 1.6066 & 79.80 & 1.6202 & 1.36 & 50 & 1.2574 & 85.17 & 1.2656 & 1.20 \\
4  & 200 & 50 & 1.6746 & 82.20 & 1.6863 & 1.49 & 50 & 1.2213 & 85.07 & 1.2286 & 1.16 \\
5  & 399 & 50 & 1.6989 & 81.77 & 1.7175 & 1.54 & 50 & 1.1515 & 82.79 & 1.1615 & 1.13 \\
6  & 399 & 50 & 1.7177 & 82.36 & 1.7352 & 1.56 & 50 & 1.1150 & 81.84 & 1.1250 & 1.11 \\
8  & 200 & 50 & 1.7584 & 83.94 & 1.7737 & 1.61 & 50 & 1.1169 & 82.16 & 1.1260 & 1.14 \\
9  & 200 & 50 & 1.7881 & 84.51 & 1.8039 & 1.65 & 50 & 1.1172 & 82.54 & 1.1245 & 1.15 \\
10 & 399 & 50 & 1.8315 & 85.43 & 1.8474 & 1.72 & 50 & 1.1167 & 83.95 & 1.1219 & 1.14 \\
\bottomrule
\end{tabular}}
\end{table}

\section{Directional Generalization Intervals and Jacobian SVD Analysis}
\label{sec:jacobian-svd}

Our experiment examines how Generalization Intervals (GIs) along Jacobian RSV directions evolve during training. The Singular Value Decomposition (SVD) decomposes the Jacobian matrix into orthogonal components, where singular values represent the network's sensitivity along corresponding input directions. For each model (early-trained after 1 epoch and later-trained after 50 epochs), we computed Jacobians at 50 correctly classified MNIST samples, performed SVD, and measured GIs along each RSV direction with maximum perturbation of 20.0 and step size of 0.01.

The results reveal a significant transformation in directional sensitivity. Early training (epoch 1) shows a relatively uniform GI profile ($\sim$7) across all RSV directions, whereas after extended training (epoch 50), GIs decrease along directions with large singular values (low-indexed RSVs) while dramatically increasing along directions with small singular values (high-indexed RSVs), reaching an average of $\sim$11.5. This pattern supports the theory that neural networks develop a natural split between an ``information space'' (directions with large singular values) where the model remains sensitive, and a ``nuisance space'' (directions with small singular values) where the model becomes increasingly invariant.

\section{Linear Approximation Analysis Along Singular Vector Directions}
\label{app:linear-approx}

We analyze the local linear behavior of neural networks by measuring how far a linear approximation remains valid along different input directions. For each RSV direction from the Jacobian SVD, we compute the maximum perturbation magnitude $\alpha$ at which the cosine similarity between the true network output and its linear approximation remains above 0.95. This threshold represents the distance over which the linear Taylor expansion of the network function serves as a reliable proxy for the true nonlinear function.

Our analysis contrasts untrained and trained MLP models on MNIST, using 50 samples (5 per class) for reliable statistics. For each sample, we compute the Jacobian at the penultimate layer, perform SVD, and measure $\alpha$ values along 100 evenly distributed RSV directions. The results reveal a striking transformation in local linearity through training. The untrained model (random initialization) exhibits uniformly poor linear approximations, with maximum $\alpha$ ranging only from 0.2-0.4 across all directions. In contrast, the trained model develops a clear monotonic relationship: directions with higher singular values (lower RSV indices) support shorter linear approximations ($\alpha \approx 0.4$), while directions with lower singular values (higher RSV indices) maintain linearity over much greater distances ($\alpha \approx 1.2$). The transition around indices 300-400, where the singular value spectrum flattens, corresponds to where trained models achieve significantly better linearity than untrained ones.

\subsection{Remark: Theorem~\ref{thm: lin-appx-gi} and the Residual Bound}
\label{app:taylor_remark}

Theorem~\ref{thm: lin-appx-gi} may be interpreted as requiring the network $f$ is approximately linear, although approximate linearity is not required for the large, non-adversarial displacements involved in test-time classification. A residual bound is assumed. The theorem follows from the exact identity $f(x_0+td) = f(x_0) + tJ_f(x_0)d + R(t)$, which holds for any differentiable $f$ without requiring the network be linear; nonlinearity is fully absorbed into the residual $R(t)$, bounded by $\delta$ over the ray. A larger nonlinear residual requires a correspondingly larger $\delta$, which reduces the magnitude of the resulting GI lower bound; it does not invalidate the theorem. Correspondingly, all GI and ray-coverage values reported in this paper are obtained through direct evaluation of the nonlinear network along each ray, without substituting a linearized network model.

Empirically, a first-order approximation is found to represent the network's behavior over extended distances along directions associated with the largest GIs. This relationship is examined through representation analysis reported in this section. The analysis above shows that the distance $\alpha$ over which a linear approximation of the penultimate-layer representation $\phi$ maintains cosine similarity $\geq 0.95$ with the true representation grows from $\alpha\approx0.2$--$0.4$ (untrained, roughly uniform across directions) to $\alpha\approx1.2$ after training, specifically along the low-curvature directions with the largest GIs. Since $f = W\phi$ for a fixed linear classifier head $W$, a small residual in $\phi$ implies a residual in $f$ scaled by $\|W\|_{\mathrm{op}}$ (the same step used in the proof of Theorem~\ref{thm:gi_spectral}), relating these observations about $\phi$ transfers to the quantity Theorem~\ref{thm: lin-appx-gi} explicitly bounds.

\section{Test-to-Train Generalization Interval Experimental Protocol}
\label{app:test-to-train-protocol}
Our ray coverage analysis examined whether test samples consistently fall within the GI zones of training samples. For each dataset, we computed:

\begin{equation}
\label{eq:coverage}
\text{Coverage}(x_{\text{test}}) = \frac{1}{|S_p|} \sum_{x_i \in S_p} {\textbf{1}}[\text{GI}(f, x_i, x_{\text{test}} - x_i) \geq 1]
\end{equation}

where $S_p$ is the set of training samples from the predicted class, and $\textbf{1}[\cdot]$ is the indicator function. This measures the percentage of training samples whose ray coverage contains the test sample.

Dataset-specific protocols were as follows:

\begin{itemize}
    \item \textbf{MNIST:} All 10,000 test samples compared against all training samples of the predicted class ($\sim$6,000 per class), using MLP, CNN, and ViT models with step size $\delta=0.01$.

    \item \textbf{CIFAR-10:} 50 test samples per class compared against 200 randomly sampled training samples of the predicted class, using ResNet20 and VGG16-BN with step size $\delta=0.01$.

    \item \textbf{ImageNette:} 20 test samples per class compared against 100 randomly sampled training samples of the predicted class, using VGG16, ResNet50, ViT-B, and DeiT-S with step size $\delta=0.01$.
\end{itemize}

For computational efficiency, we implemented batch processing tailored to each dataset's dimensionality (6,000/2,000/100 sample pairs per batch for MNIST/CIFAR-10/ImageNette respectively). The consistently high coverage percentages across datasets support our Neural Expansion hypothesis, confirming that classification operates through ray coverage rather than instance-level memorization.

\section{Robustness and Discriminative Validity of Ray Coverage}
\label{app:rc_validity}

The ray-coverage measurements used throughout the paper are evaluated with respect to training variability and class specificity. Stability across sources of training randomness is examined, and class-specific structure is assessed through comparisons involving untrained networks and rays directed toward different classes.

\subsection{Discriminative Baselines: Untrained Networks and Wrong-Class Rays}
\label{app:rc_discriminative}

A 2$\times$2 study was conducted (untrained vs.\ trained checkpoint $\times$ same-class vs.\ different-class rays) on MLP-MNIST, CNN-MNIST, and CNN-Fashion-MNIST, using $n=100$ test samples per architecture and the step-search protocol of Appendix~\ref{app:analysis-methods}. For the different-class condition, the correct class is held fixed at the test sample's predicted class $c$, while rays are evaluated toward 3 random wrong-class training pools per test sample; this ensures a consistent comparison, since directly reusing the per-class coverage columns from Appendix~\ref{app:test-to-train-protocol} is invalid (it makes GI $\leq 1$ tautological, and empirically yields exactly 0\%).

\begin{table}[ht]
\centering
\caption{Ray coverage (\%) under a 2$\times$2 untrained/trained $\times$ same/different-class design. Training substantially increases same-class coverage, and trained same-class coverage substantially exceeds trained different-class coverage, confirming coverage is class-discriminative rather than an artifact of the measurement.}
\label{tab:rc_discriminative}
\begin{tabular}{lcccc}
\toprule
Model & Untrained, same & Untrained, diff. & Trained, same & Trained, diff. \\
\midrule
MLP, MNIST & 46.8\% & 30.6\% & 88.9\% & 1.2\% \\
CNN, MNIST & 35.6\% & 42.7\% & 96.3\% & 0.3\% \\
CNN, Fashion-MNIST & 32.3\% & 34.6\% & 88.5\% & 1.0\% \\
\bottomrule
\end{tabular}
\end{table}

At $p<10^{-11}$, training substantially increases same-class coverage (32--47\%$\to$89--96\%), and trained same-class coverage substantially exceeds trained different-class coverage (89--96\% vs.\ 0.3--1.2\%). The untrained same-vs-different contrast is nonsignificant on Fashion-MNIST ($p=0.19$) but statistically significant on MNIST ($p=3.9\times10^{-12}$ MLP, $p=0.02$ CNN), indicating a mild same-class pixel bias in untrained networks, with a substantially larger class-specific effect being observed following network training.

\subsection{Statistical Robustness Across Seeds and Optimizers}
\label{app:rc_stat_robustness}

The same-class ray-coverage measurement was also repeated (MNIST MLP/CNN, percentage of pairs with GI $>1$, protocol as in Appendix~\ref{app:test-to-train-protocol}) across 5 random seeds and 3 optimizers (SGD, Adam, RMSprop; learning rate 0.001) for both architectures, 16 models total, recomputing coverage at several checkpoints from untrained through epoch 20 on a fixed $n=50$ evaluation split.

\begin{table}[ht]
\centering
\caption{Final-epoch ray coverage (\%) is nearly seed-invariant and only mildly optimizer-sensitive, whereas substantial variability is observed among untrained baselines (13--99.96\% MLP, 24--95\% CNN).}
\label{tab:rc_stat_robustness}
\begin{tabular}{llccccc}
\toprule
Architecture & Factor & $n$ runs & Mean & Std & CV (\%) & Min--max \\
\midrule
MNIST MLP & Seed & 5 & 88.91 & 0.07 & 0.1 & 88.82--89.02 \\
MNIST MLP & Optimizer & 3 & 91.20 & 1.64 & 1.8 & 89.04--93.03 \\
MNIST CNN & Seed & 5 & 96.86 & 0.11 & 0.1 & 96.74--97.04 \\
MNIST CNN & Optimizer & 3 & 98.52 & 1.22 & 1.2 & 96.79--99.43 \\
\bottomrule
\end{tabular}
\end{table}

The untrained-vs-final difference is significant ($p<0.05$, Mann-Whitney U) in 15 of 16 runs, most at $p<10^{-14}$; the one exception (CNN seed 2, $p=0.061$) exhibits the same trend, with elevated coverage already observed at the untrained checkpoint for this seed (95.5\%) at initialization. The untrained baseline is itself seed-dependent, while the trained endpoint converges tightly regardless of seed or optimizer (CV $\leq 1.8\%$), supporting the stability of coverage measurements used for the results in the main text.

\subsection{Orthogonal Tube Test: Full Protocol and Results}
\label{app:tube_test}

A class-preserving straight-line ray from a training sample to a test sample establishes stability only along that one direction; the extent of prediction stability in the surrounding neighborhood must therefore be assessed through additional directional measurements. This property is evaluated using an orthogonal ``tube'' probe: for 100 previously identified rays (GI $>1$) per model on a trained MNIST-CNN and a trained Fashion-MNIST-CNN, we sample 20 directions exactly orthogonal to each ray at 5 evenly spaced points along it, apply the established step-search algorithm (Appendix~\ref{app:analysis-methods}) along directions perpendicular to the ray, and record the distance to the first class change (the \emph{tube radius}), normalized by the ray's own hop distance $\|x_{\text{test}}-x_i\|$. CNNs are specifically selected, whose local receptive fields make an orthogonal perturbation more likely to induce an early transition than in a fully-connected network, as a challenging case for this test.

\begin{table}[ht]
\centering
\caption{Orthogonal tube test: orthogonal stability radius around high-coverage rays, normalized by the ray's own hop distance. A ratio $\geq 1$ indicates the stable region is at least as wide as the ray is long, in every perpendicular direction tested.}
\label{tab:tube_test}
\begin{tabular}{lccc}
\toprule
Model & Median hop distance & Median tube radius & Median ratio (tube/hop) \\
\midrule
MNIST CNN & 17.9 & 23.1 & 1.39$\times$ \\
Fashion-MNIST CNN & 16.2 & 15.4 & 1.01$\times$ \\
\bottomrule
\end{tabular}
\end{table}

On both CNNs, the typical high-coverage ray lies within a region at least as wide as the ray itself (Table~\ref{tab:tube_test}), across every one of the 20 perpendicular directions sampled at every one of the 5 points along the ray. Ray coverage therefore reflects an extended multidimensional same-class region, not a one-dimensional artifact of the specific train-test direction measured.

\section{Class Region Analysis Along Train-to-Test Paths}
\label{app:class_region_analysis}
\begin{figure}[ht]
    \centering
    \begin{subfigure}[b]{0.48\linewidth}
        \includegraphics[width=\linewidth]{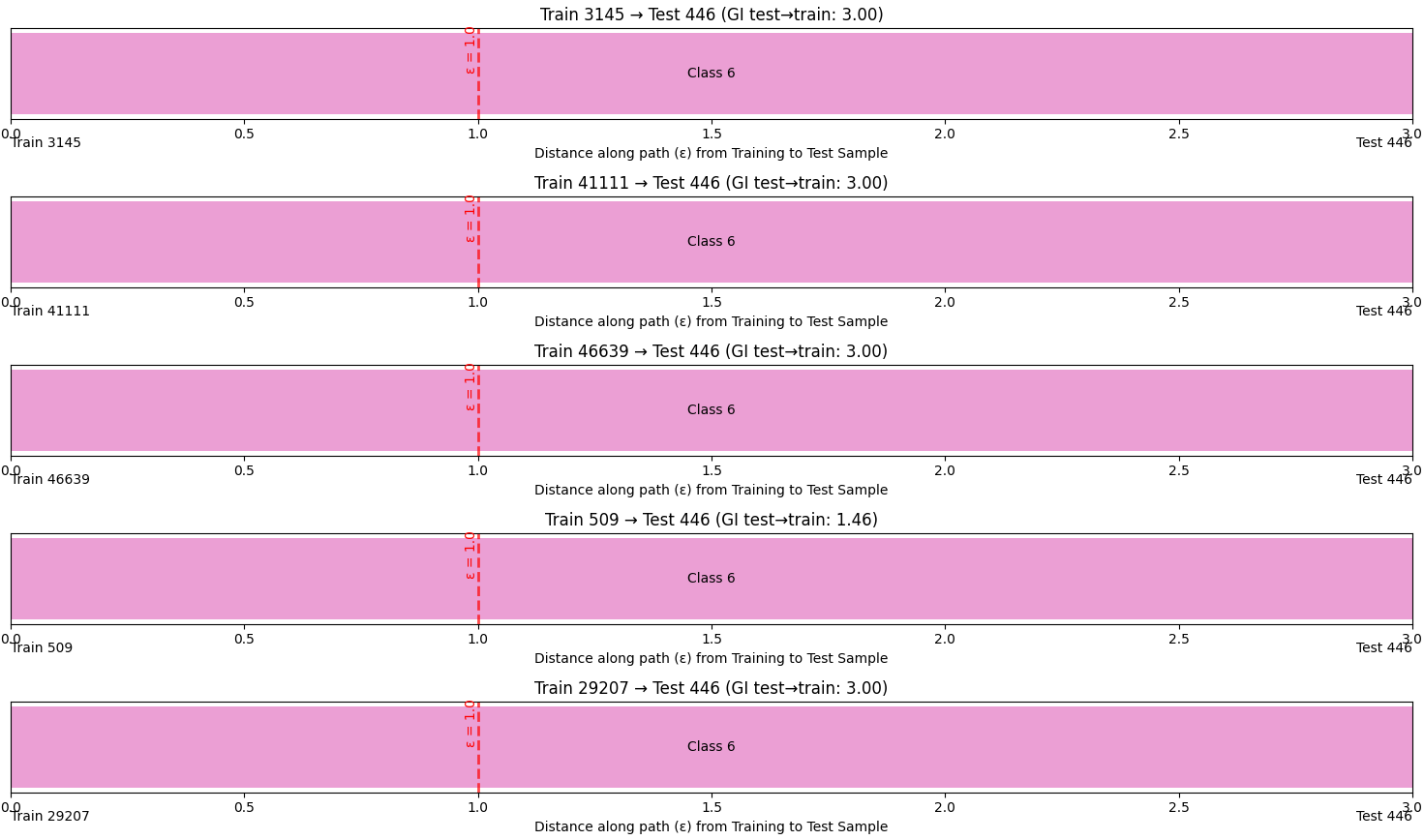}
        \caption{}
    \end{subfigure}
    \hfill
    \begin{subfigure}[b]{0.48\linewidth}
        \includegraphics[width=\linewidth]{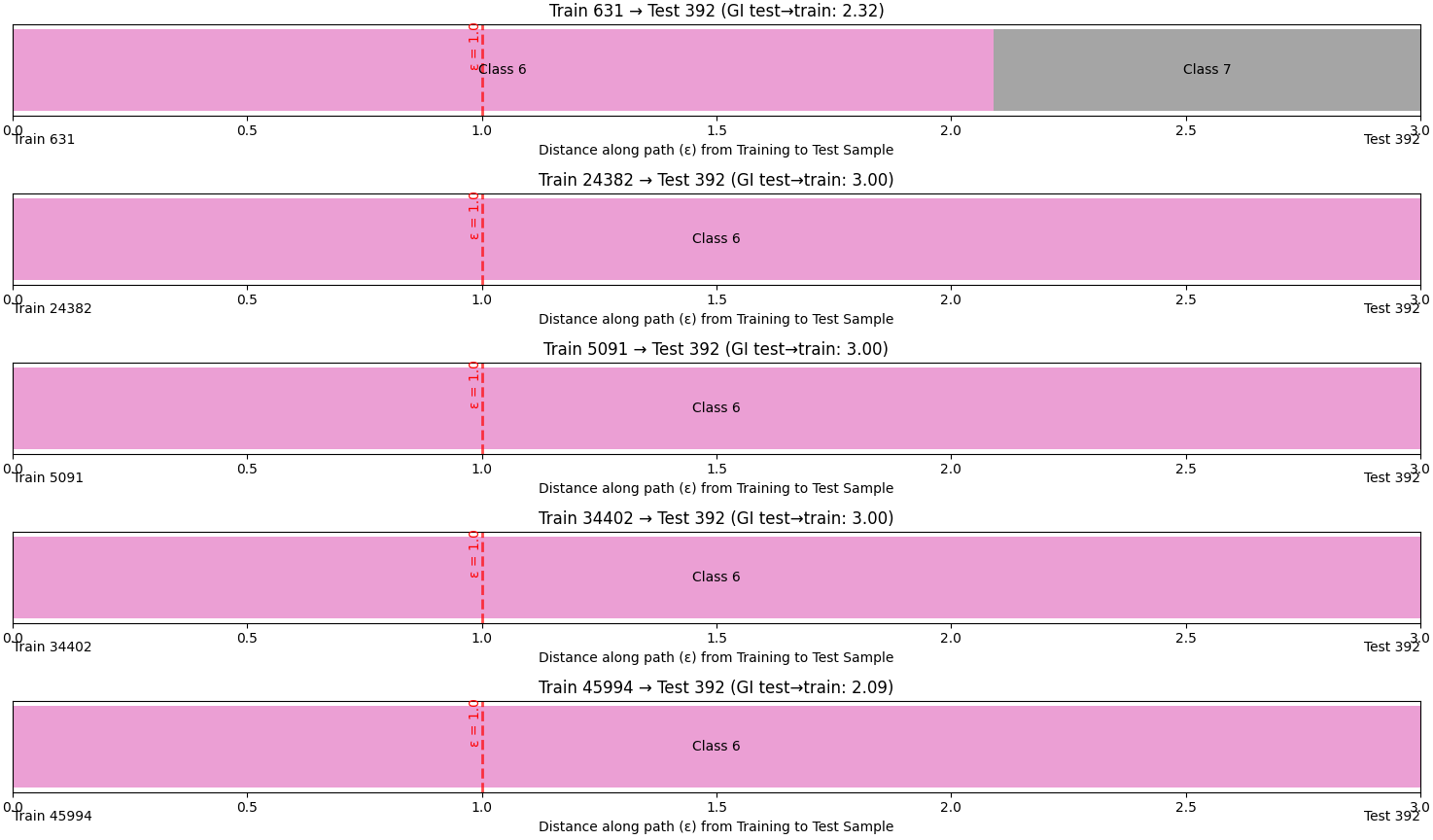}
        \caption{}
    \end{subfigure}
    \caption{Predicted class trajectories along paths from training to test samples (Train $\to$ Test Class 6) with GI $>$ 1.0. Each subplot shows class-stable behavior across the entire interpolation path, indicating the test point lies within the ray coverage of the training sample.}
    \label{fig:gi_gt1_examples}
\end{figure}

This section investigates how a model's class predictions evolve along linear paths from training samples to test samples of the same class. We conduct this analysis using a Multi-Layer Perceptron (MLP) trained on the MNIST dataset. We define $\varepsilon \in [0, \varepsilon_{\text{max}}]$ as a scalar interpolation parameter that traces a linear trajectory from a starting input $x_1$ to a target input $x_2$ in input space. Each point along the path is given by $x(\varepsilon) = x_1 + \varepsilon (x_2 - x_1)$. The generalization interval (GI) is the maximum value of $\varepsilon$ for which the model's predicted class remains consistent with the initial prediction at $x_1$.
We group path analyses into two categories:

\textbf{GI $>$ 1.0:} The test sample lies within the ray coverage of the training point. These paths typically show a single, uninterrupted class prediction (matching the true class) along the entire trajectory (see Figure~\ref{fig:gi_gt1_examples}).

\begin{figure}[ht]
    \centering
    \begin{subfigure}[b]{0.48\linewidth}
        \includegraphics[width=\linewidth]{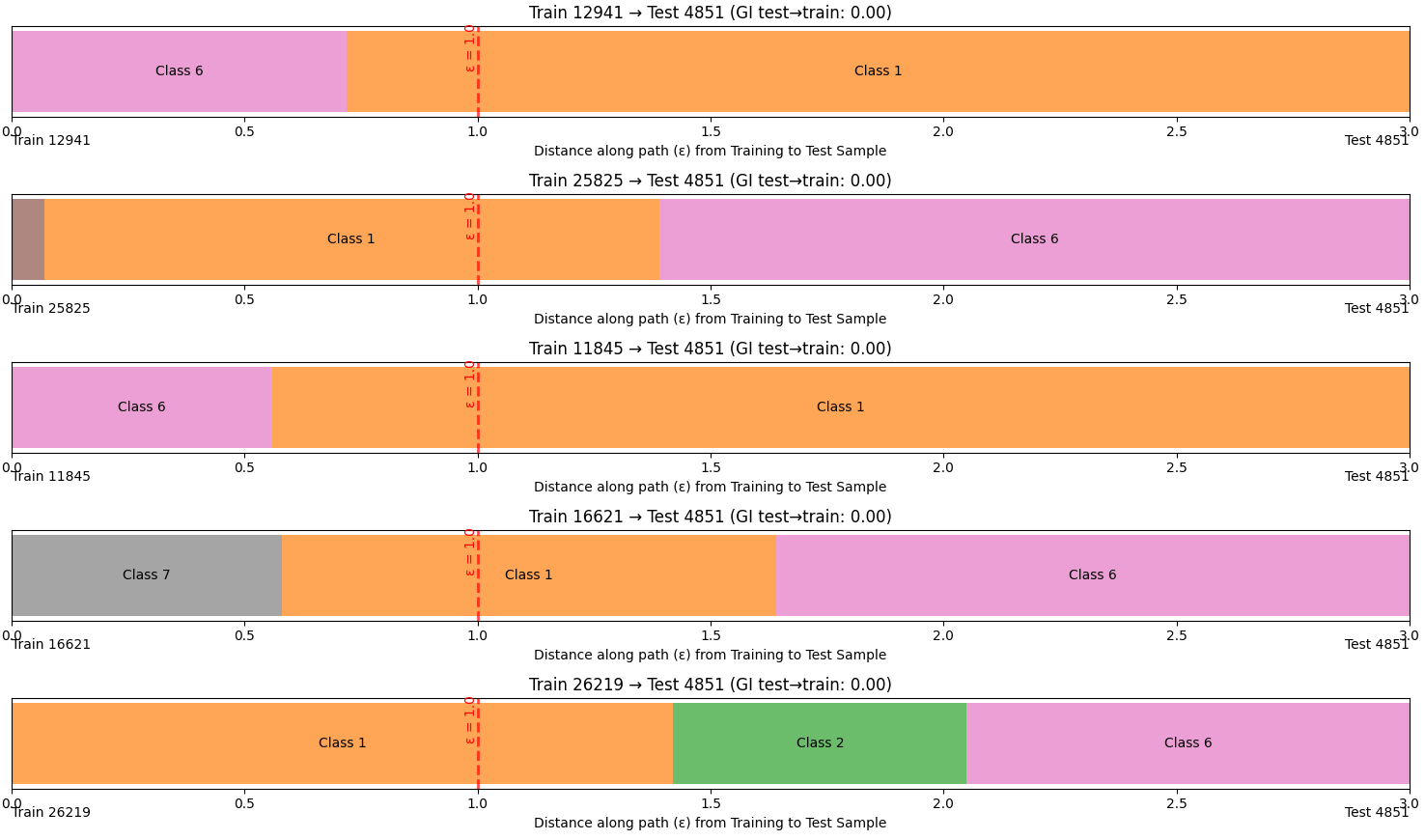}
        \caption{}
    \end{subfigure}
    \hfill
    \begin{subfigure}[b]{0.48\linewidth}
        \includegraphics[width=\linewidth]{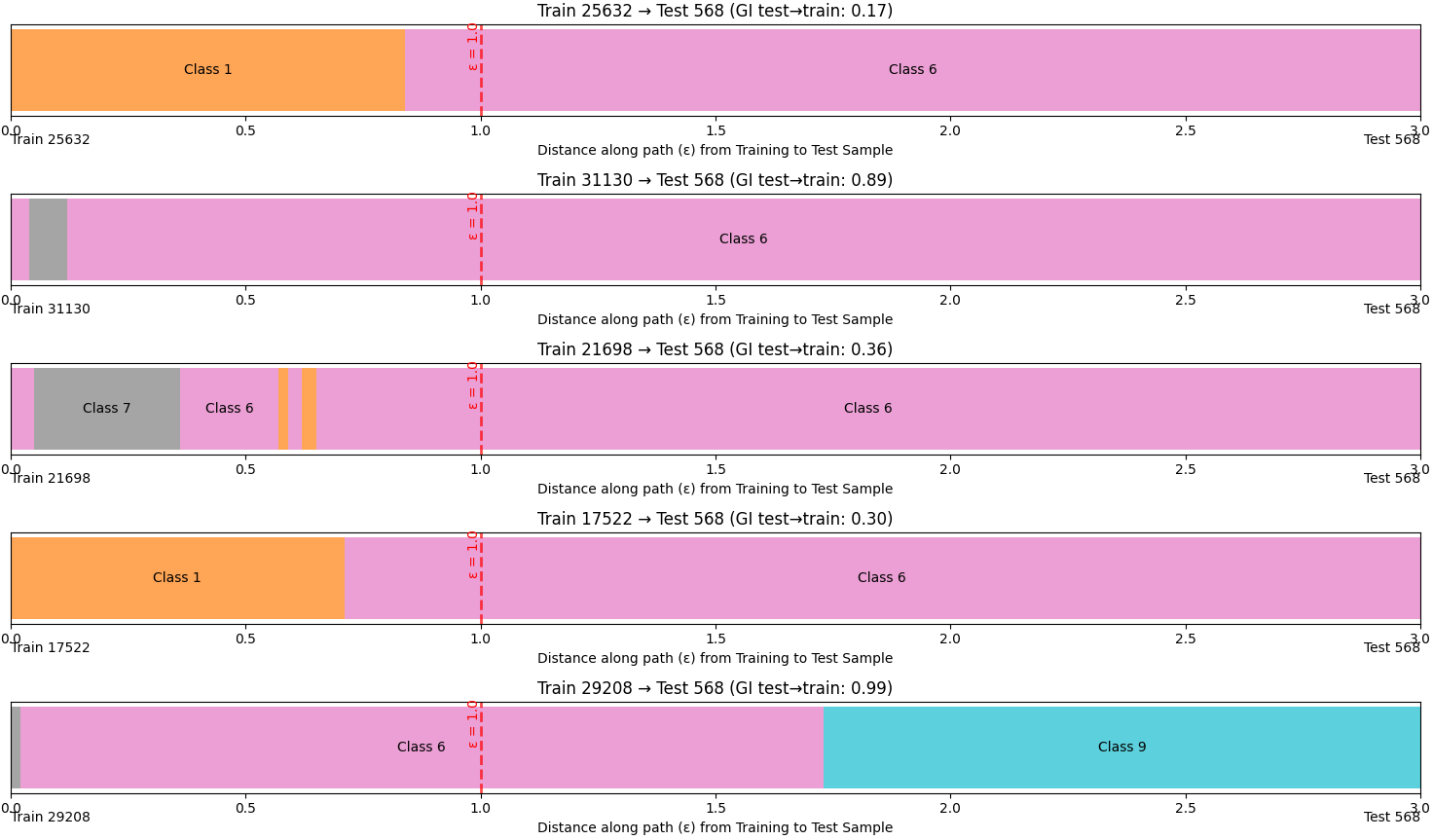}
        \caption{}
    \end{subfigure}
    \caption{Predicted class trajectories along paths from training to test samples (Train $\to$ Test Class 6) with GI $<$ 1.0. Multiple class transitions are observed along the interpolation, indicating unstable or fragmented class regions.}
    \label{fig:gi_lt1_examples}
\end{figure}

\textbf{GI $<$ 1.0:} The test sample lies outside the ray coverage region. These paths frequently exhibit multiple class transitions, indicating that the model's classification is unstable along the direction from train to test (see Figure~\ref{fig:gi_lt1_examples}).

Together, these visualizations highlight a key behavioral difference: when GI $>$ 1.0, the classification is stable and coherent between the training and test sample. In contrast, GI $<$ 1.0 is associated with fragmented decision regions and unstable class behavior, even between samples of the same true class.

\section{Causal Intervention Experiment: Full Protocol}
\label{app:causal_intervention}

\subsection{Derivation of the Population Error Identity}
\label{app:causal_derivation}
We  consider binary classification task with labels $0, 1$ and fix the training set $S= \{(x_i, y_i)\}_{i=1}^{N}$ sampled from the population. We denote a random sample from the population by $(X, Y) \in \mathbb{R}^d\times \mathbb{R}$. For each sample $(X,Y)$ from the population, we assign a sample from the training set $(x_I,y_I)$ so that $y_I=Y$. The random index $I$ keeps track of which GI to keep track of at sample $(X,Y)$, since GIs from multiple samples may reach the sample $(X,Y)$. It is just a device to make sure that there is no ambiguity about which GI we are considering, and the choice is generally interchangeable. We now define, for $B\subset \mathbb{S}^{d-1}$, where $\mathbb{S}^{d-1}$ is the $d-1$-dimensional sphere in $\mathbb{R}^d$,
\begin{equation}
    \mu_i(B) = P\left(I=i,\quad  \frac{X-x_I}{\|X-x_I\|}\in B \right)
\end{equation}
This is the probability that a random sample $X$ assigned to $x_i$ points to the set of directions $B$ when we see $x_i$ as the origin. In other words, this tells us which directions the population is most concentrated in, when seen from $x_i$. We assume that the random variable $R=\|X-x_I\|_2$ admits a density conditional to the direction $U=\frac{X-x_I}{\|X-x_I\|}$, denoted $p_i(r|u)$:

\begin{equation}
    P(I=i,U\in du,R\in dr)=p_i(r|u)dr\mu_i(du)
\end{equation}

Now, let $f_n:\mathbb{R}^n\to \{0,1\}$ denote the prediction model at training time $n$. $f_n(x)$ denotes the predicted class of $x$ by the model. One nontrivial assumption we make is the following \emph{no-re-entry assumption}:
\begin{equation}
    f_n(x_I+RU)=y_I \Leftrightarrow R<\mathrm{GI}(f_n,x_I,U), \quad \textrm{almost surely.}
\end{equation}
This statement says that, along the ray from the assigned sample $x_I$, the model predicts the correct original class $y_I$ up to the GI boundary and not beyond. Thus, the prediction does not "re-enter" the class $y_I$. We note that this does not mean samples of class $y_I$ do not occur beyond GI. In such a case, the model misclassifies these samples.

Now, we will consider the population error of the model: $\epsilon_n = P(f_n(X)\neq Y)$. We note that the statement $f_n(x_I+RU)=y_I$ is equivalent to $f_n(X)= Y$ by definitions of $I, R, U$, so the no-entry assumption means that
\begin{align}
    \epsilon_n &=P(R\geq\mathrm{GI}(f_n,x_I,U)) \\ &= \sum_{i}\int_{\mathbb{S}^{d-1}}\int_{\mathrm{GI}(f_n,x_i,u)}^{\infty}p_i(r|u)dr\mu_i(du)
\end{align}
Therefore, the difference in error between $f_{n+1}$ and $f_n$ is

\begin{equation}
\label{eqn:error-difference}
    \epsilon_{n+1} - \epsilon_n = -\sum_{i}\int_{\mathbb{S}^{d-1}}\int_{\mathrm{GI}(f_n,x_i,u)}^{\mathrm{GI}(f_{n+1},x_i,u)} p_i(r|u)dr\mu_i(du)
\end{equation}
Remarkably, if the GI increased, i.e., $\mathrm{GI}(f_n,x_i,u)\leq \mathrm{GI}(f_{n+1},x_i,u)$ almost surely, then the right-hand side of \eqref{eqn:error-difference} is nonpositive, so the error decreases (or does not change). We remark that without the no-re-entry assumption, the conclusion is generally false. For example, for the population distribution of $X\sim \mathrm{Unif}([0,3])$ and $Y=+1$, the single training example of $(x_1,y_1)=(0,1)$, and a training sequence of $f_0(x)=\textbf{1}{\{(x-1)(x-2)>0\}}$ and $f_1(x)=\textbf{1}{\{1-x>0\}}$ , the GI at $x_1$ to the positive direction is unchanged at $1$, but the population error increases because the interval $[2,3]$, which was correctly classified, turns into incorrectly classified interval, while not affecting GI because $[2,3]$ is disconnected with the region covered by GI, i.e., $[0,1]$.

We will complement this theoretical observation with the experiment below.

\subsection{Interventional Experiment: Design}
\label{app:causal_design}

For both the MNIST MLP (784--1024--512--10, Appendix~\ref{app:model_architectures}) and a two-block convolutional Fashion-MNIST CNN with BatchNorm, an auxiliary loss term is added, active in two of three training arms, that acts directly on what a GI measures: whether the prediction remains unchanged under local perturbations around a training point $x$. For each $(x,y)$ we draw a random unit direction $u$ and perturb by $\varepsilon(x) = 0.08\times\|x\|_2$, placing the step in the 2.0--2.2 range in this normalized space, comparable to standard $\ell_2$ adversarial-perturbation budgets for MNIST-scale data: a local perturbation that is applied independently of another sample location. The hinge objective uses the margin at perturbed inputs, defined as the model's output for the true class minus its highest output among the other classes:
\begin{itemize}
    \item \textbf{Expand} arm: $\mathcal{L} = \text{CE} + \lambda\cdot\text{ReLU}(\tau - \text{margin}(x+\varepsilon u))$, encouraging confident correct classification of the perturbed input.
    \item \textbf{Restrict} arm: $\mathcal{L} = \text{CE} + \lambda\cdot\text{ReLU}(\text{margin}(x+\varepsilon u) + \tau)$, encouraging confident misclassification of perturbed inputs, while cross-entropy encourages correct classification at $x$.
    \item \textbf{Control} arm: standard cross-entropy, $\lambda=0$.
\end{itemize}
We set $\tau=2.5$ to match the mean margin measured on our own trained checkpoint (2.48 over 1000 training images), a target calibrated using observations. At every epoch we measure GI and ray coverage with our existing, unmodified step-search routine (Appendix~\ref{app:analysis-methods}), on a fixed pool of 200 random same-class train-to-train pairs, independent of the training seed and identical across arms. Each configuration is trained with 3 seeds, reported both individually and averaged.

\subsection{Interventional Experiment: Full Results}
\label{app:causal_results}

Table~\ref{tab:causal_intervention} in the main text reports the final-epoch (epoch 10) results averaged over 3 seeds. Individually, restricting ray coverage on MNIST-MLP lowers test accuracy to 74.7\%, 76.5\%, and 74.3\% across the 3 seeds ($p=0.003$ vs.\ control), while expansion yields consistently higher accuracy across seeds (std 0.08, $p<10^{-4}$). On Fashion-MNIST-CNN, restriction produces a statistically significant reduction ($-7.4$pp accuracy, $-19.5$pp ray coverage, $p\approx0.008$ on both); expanding increases test accuracy by 1.1pp ($p=0.008$), with a smaller corresponding increase in ray coverage on this architecture. In both datasets, GI and ray coverage separate in the intended direction from epoch 1 onward, on every individual seed, with no overlap between arms. The reported MNIST-MLP restrict arm's train accuracy (73.9\%) is nearly as low as its test accuracy (75.2\%), indicating the effect is a slower formation of classification structure overall, rather than merely a widened train/test gap.

\section{Ray Coverage in Challenging Cases}
\label{app:ray_cover_challenging}

\subsection{Random Label Experiment Protocol}
\label{app:random_label}

We evaluate ray coverage in two randomized settings:
\begin{itemize}
    \item \textbf{RANDOM-10}: Training labels are reassigned within 10 classes.
    \item \textbf{RANDOM-100}: Both training and test labels are reassigned across 100 classes.
\end{itemize}

Despite the lack of semantic label structure, ray coverage remains active in both settings. Using an MLP architecture (\(784\)-\(1024\)-\(512\)-\(n\), where \(n\) is 10 or 100) with a GI step size of \(\delta = 0.01\), we evaluated all 10,000 test samples against all training samples of the predicted class.

\begin{table}[ht]
\centering
\caption{Ray coverage percentages for correctly and incorrectly classified test samples under random label supervision.}
\label{tab:random_labels}
\begin{tabular}{lcc}
\toprule
\textbf{Setting} & \textbf{Correctly Classified} & \textbf{Misclassified} \\
\midrule
RANDOM-10 (MNIST-MLP) & 17\% & 24\% \\
RANDOM-100 (MNIST-MLP) & 18\% & 22\% \\
\bottomrule
\end{tabular}
\end{table}

Although label randomization reduced ray coverage (17\%-24\% on average) as shown in Table~\ref{tab:random_labels}, it remained the classification mechanism, in contrast to the higher coverage seen with semantically meaningful labels.

\subsection{Long-tailed Distribution Analysis Details}
\label{app:long_tail_distribution}
We analyze neural network behavior under two long-tailed MNIST variants: \textbf{MNIST-LT1} and \textbf{MNIST-LT2}, which differ in the severity of class imbalance (6.5:1 and 78.6:1, respectively) (see details in Table~\ref{tab:mnist_distribution}). In \textbf{MNIST-LT2}, which has a more extreme imbalance, we observe systematic migration of misclassified samples. As shown in Figure~\ref{fig:random_longtail}(b), samples from low-frequency classes tend to be absorbed into the ray coverage of higher-frequency classes.

Despite the imbalance, ray coverage remains strong across all class groups, as shown in Figure~\ref{fig:random_longtail}(c), but the ray coverage from high-frequency classes appears to dominate test-time decisions. In \textbf{MNIST-LT1}, misclassified samples also showed similar migration patterns: low-frequency classes tended to migrate to medium (47.2\%) and high (35.2\%) frequencies, with only 17.6\% staying within the low-frequency group. In both settings, the GI coverage remained consistently high (86.6\%-94.3\%) across all frequency combinations, confirming that \textbf{Neural Expansion} operates reliably across varying imbalance conditions.

\begin{table}[h]
\centering
\small
\caption{Class distributions in MNIST long-tailed variants. H=High, M=Medium, L=Low frequency bands based on 33rd and 66th percentiles.}
\label{tab:mnist_distribution}
\begin{tabular}{c|ccccccccccc}
\toprule
\textbf{Class} & 0 & 1 & 2 & 3 & 4 & 5 & 6 & 7 & 8 & 9 & \textbf{Ratio} \\
\midrule
\textbf{LT1} & 4717 & 2866 & 2370 & 2015 & 1806 & 1590 & 1346 & 1034 & 886 & 730 & 6.5:1 \\
\textbf{Band} & H & H & H & M & M & M & M & L & L & L & \\
\midrule
\textbf{LT2} & 4716 & 2504 & 2026 & 1464 & 1096 & 761 & 383 & 218 & 93 & 60 & 78.6:1 \\
\textbf{Band} & H & H & H & M & M & M & L & L & L & L & \\
\bottomrule
\end{tabular}
\end{table}

\subsection{Adversarial Examples Experiment Protocol}
\label{app:adversarial_examples_experiments}
We conducted a systematic analysis of how adversarial examples interact with ray coverage across three datasets and architectures:

\begin{itemize}
   \item \textbf{MNIST-MLP}: Custom MLP with architecture 784-1024-512-10
   \item \textbf{CIFAR10-ResNet20}: Pre-trained ResNet20 from PyTorch hub
   \item \textbf{ImageNette-ViTB}: Pre-trained Vision Transformer Base (patch size 16)
\end{itemize}

For each configuration, we generated adversarial examples using three established attack methods:

\begin{itemize}
   \item \textbf{Carlini-Wagner (CW) \citep{carlini2017evaluatingrobustnessneuralnetworks}}: Optimization-based attack that minimizes perturbation magnitude with parameters $c=1e\text{-}2$, learning rate $1e\text{-}2$, and 1000 iterations.
   \item \textbf{Iterative Fast Gradient Sign Method (IFGSM) \citep{kurakin2017adversarialexamplesphysicalworld}}: Iterative gradient-based attack with $\epsilon=0.03$, $\alpha=0.005$, and 10 iterations.
   \item \textbf{Projected Gradient Descent (PGD) \citep{madry2019deeplearningmodelsresistant}}: Extension of IFGSM with random initialization, using $\epsilon=0.03$, $\alpha=0.005$, and 30 iterations.
\end{itemize}

Our dataset-specific protocols involved:
\begin{itemize}
   \item \textbf{MNIST}: 20 samples/class (total 200), compared against all training samples of predicted class
\item \textbf{CIFAR-10}: 10 samples/class (total 100), compared against 200 training samples of predicted class
\item \textbf{ImageNette}: 5 samples/class (total 50), compared against 100 training samples of predicted class\end{itemize}

\begin{itemize}
    \item \textbf{MNIST-MLP:} Correctly classified adversarial examples show 83.9--86.6\% coverage with their predicted class, while misclassified examples show 91.2--94.5\%
    \item \textbf{CIFAR10-ResNet20:} Correctly classified adversarial examples have moderate coverage (63.9--68.7\%), and misclassified examples vary between 39.6--58.1\% depending on attack strength
    \item \textbf{ImageNette-ViT-B:} Correctly classified adversarial examples exhibit low ray coverage (15.8--19.6\%). For misclassified examples, coverage could not be computed reliably due to class predictions falling outside the 10-class subset and into the full 1000-class ImageNet space
\end{itemize}

\begin{figure}[ht]
\centering
\includegraphics[width=\linewidth]{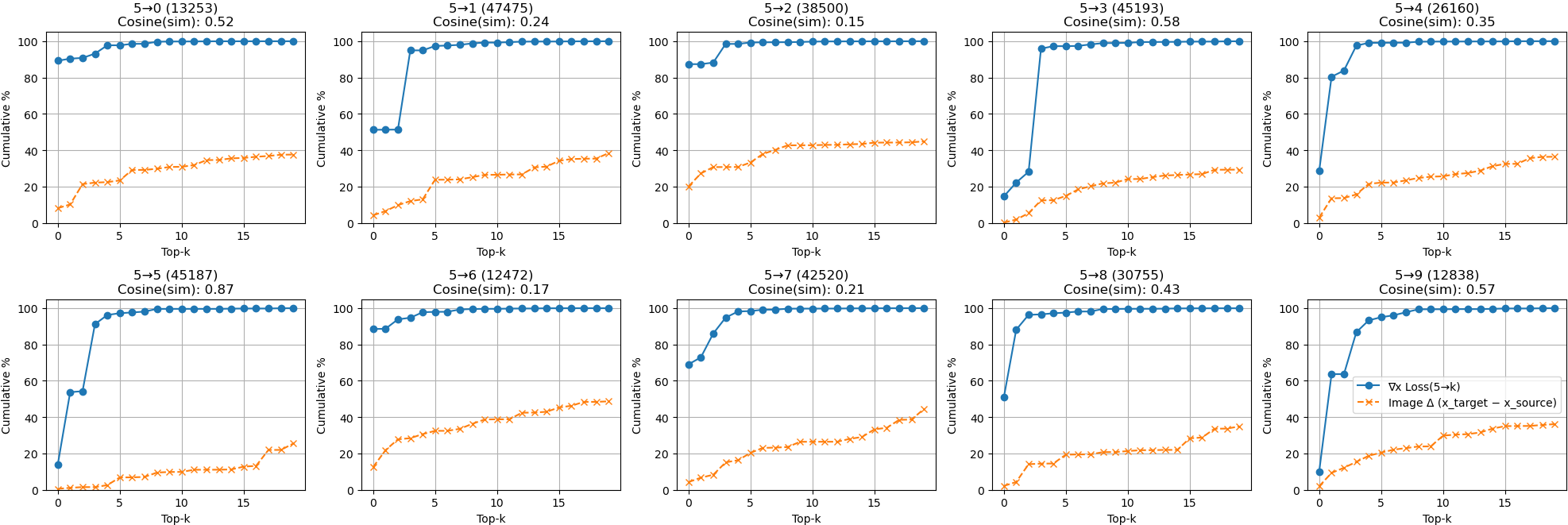}
\caption{Gradient \& Image $\Delta$ Projections on Top-k Right Singular Vectors. Blue lines show the cumulative percentage of the loss gradient projected onto the top-k RSVs. Orange lines show the projection of the image difference vector between target and source images.}
\label{fig:adv_projection}
\end{figure}

\textbf{Adversarial Alignment Analysis:} We computed the Jacobian SVD at each source image and measured the cumulative projection of both adversarial gradients and image-to-image difference vectors onto the right singular vectors (see Figure~\ref{fig:adv_projection}). Loss gradients (blue lines) concentrate heavily in the first few singular vector directions, often reaching 80-95\% cumulative projection within the top 5 directions. In contrast, natural image difference vectors (orange lines) show a more gradual accumulation, typically reaching only 20-40\% in the top 5 directions. This contrast explains why adversarial perturbations can be much smaller than natural image differences: they selectively target the model's most sensitive directions, precisely those with the shortest generalization intervals.

Our sample counts across dataset/model/attack combinations were:
MNIST-MLP: CW (132 correct, 68 wrong), IFGSM (81 correct, 19 wrong), PGD (74 correct, 26 wrong);
CIFAR10-ResNet20: CW (37 correct, 63 wrong), IFGSM (25 correct, 75 wrong), PGD (35 correct, 65 wrong);
ImageNette-ViTB: CW (53 correct), IFGSM (47 correct), PGD (42 correct).

\subsection{Out-of-Distribution Sample Analysis}
\label{app:ood_sample_analysis}

\begin{figure}[ht!]
\centering
\includegraphics[width=0.8\linewidth]{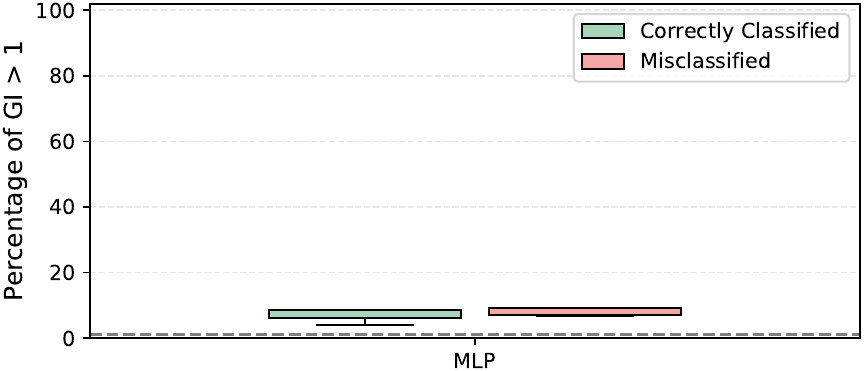}
\caption{Box plot illustrating the GI ray coverage for out-of-distribution (OOD) samples from CIFAR-10, evaluated on an MLP trained on MNIST. The gray dashed line (as defined in Figure~\ref{fig:test_samples_ray_coverage}) indicates the 1\% directional coverage threshold.}
\label{fig:gi_ood}
\end{figure}

To assess whether out-of-distribution (OOD) inputs fall within learned class-preserving regions, we compute GI ray coverage for CIFAR-10 test samples evaluated on an MLP trained on MNIST. Since the model expects $28 \times 28$ grayscale inputs, CIFAR-10 images are first transformed to grayscale $28 \times 28$ images. We follow the MNIST evaluation protocol described in Appendix~\ref{app:test-to-train-protocol}, using 50 test samples per CIFAR-10 class. Despite being OOD, CIFAR-10 samples exhibit nontrivial GI coverage of $\sim$10\%, showing that the model's learned ray coverage regions extend into OOD input space. We note that since CIFAR-10 object classes have no semantic correspondence to MNIST digit classes, the ``correct/incorrect'' stratification here reflects only whether the predicted digit index happens to match the CIFAR-10 class index, and we report aggregate coverage values accordingly.

Potential extensions of this analysis include evaluation under distribution shift (e.g., ImageNet$\to$ImageNet-Sketch/-R), self-supervised representations (e.g., DINO or CLIP linear probes), and next-token prediction in language models, where the same GI and ray-coverage formalism applies directly to context-level prediction stability.

\section{Proof of Theorem \ref{thm: lin-appx-gi}}
\label{sec: proof-thm}
By definition of the linear approximation, we have $f(x_0 + \epsilon d) = f(x_0) + \epsilon J_f(x_0)d + R$. In particular, for each component, we have $f^i(x_0+\epsilon d)  = f^i(x_0) + \epsilon [J_f(x_0)d]^i + R^i$. If $\epsilon = \textrm{GI}(f, x_0, d)$, then since we are at the decision boundary, we must have $f^y(x_0 +\epsilon d) = f^i(x_0 + \epsilon d)$ where $y$ is the class of $x_0$ and $i$ is another class.  Substituting the linear approximation, we obtain
\begin{equation}
    f^y(x_0) + \epsilon [J_f(x_0)d]^y + R^y =f^i(x_0) + \epsilon [J_f(x_0)d]^i + R^i
\end{equation}
Reorganizing the equation, we obtain
\begin{equation}
    \epsilon ([J_f(x_0)d]^y - [J_f(x_0)d]^i) =f^i(x_0) - f^y(x_0)  + R^i - R^y
\end{equation}
By the inhomogeneous assumption, $[J_f(x_0)d]^y - [J_f(x_0)d]^i \neq 0$ and thus
\begin{equation}
    \epsilon  =\frac{f^i(x_0) - f^y(x_0)  + R^i - R^y}{[J_f(x_0)d]^y - [J_f(x_0)d]^i}
\end{equation}
By the triangle inequality,
\begin{align}
    |\epsilon| = \epsilon \geq \frac{|f^i(x_0) - f^y(x_0)|  - |R^i - R^y|}{|[J_f(x_0)d]^y| + |[J_f(x_0)d]^i|}
\end{align}
Note that, by the triangle inequality again,
\begin{align}
    |R^i - R^y| \leq \sum_{j=1}^c|R^j| = \|R\|_1
\end{align}
and similarly $|[J_f(x_0)d]^y| + |[J_f(x_0)d]^i| \leq \|J_f(x_0)d\|_1$. Thus we have
\begin{equation}
    \epsilon \geq \frac{|f^i(x_0) - f^y(x_0)|  - \|R\|_1}{\|J_f(x_0)d\|_1}
\end{equation}
We note that we must have $f^y(x_0) \geq f^i(x_0)$ so that $x_0$ is classified as $y$. Thus
\begin{equation}
    \epsilon \geq \frac{f^y(x_0) - f^i(x_0)  - \|R\|_1}{\|J_f(x_0)d\|_1}
\end{equation}
Here $i=i_*$ denotes the particular wrong class whose logit ties $f^y$ at the point $x_0+\epsilon d$ where the decision boundary is first reached. Since $i_*\neq y$ is one of the $c-1$ wrong classes, we have $f^{i_*}(x_0) \leq \textrm{max}_{i\neq y}f^i(x_0)$, and therefore
\begin{equation}
    f^y(x_0) - f^{i_*}(x_0) \;\geq\; f^y(x_0) - \textrm{max}_{i\neq y}f^i(x_0) \;=\; q_0.
\end{equation}
Since the margin is $q_0 = f^y(x_0) - \textrm{max}_{i\neq y}f^i(x_0)$ and also $\|R\|_1<\delta$, we have
\begin{equation}
    \epsilon \geq \frac{q_0  - \delta}{\|J_f(x_0)d\|_1}
\end{equation}

\section{Proof of Theorem~\ref{thm:gi_spectral}}
\label{app:proof_thm_gi_spectral}

We establish the bound
$\mathrm{GI}(f, x_0, v_i) \geq \frac{q_0 - \delta}{\sqrt{c}\,\|W\|_{\mathrm{op}}\sigma_i}$
by starting from Theorem~\ref{thm: lin-appx-gi} and evaluating its denominator
explicitly at $d = v_i$ using the SVD structure of $J_\phi(x_0)$.

\paragraph{Step 1: Apply Theorem~\ref{thm: lin-appx-gi} at $d = v_i$.}

Setting $d = v_i$ (a unit vector by definition of the SVD) in
Theorem~\ref{thm: lin-appx-gi}:
\begin{equation}
    \mathrm{GI}(f, x_0, v_i) \;\geq\;
    \frac{q_0 - \delta}{\|J_f(x_0)\,v_i\|_1}
\end{equation}
It suffices to upper-bound $\|J_f(x_0)\,v_i\|_1$, since a larger denominator
yields a smaller (still valid) lower bound on GI.

\paragraph{Step 2: Convert $\ell^1$ to $\ell^2$ norm.}

For any $z \in \mathbb{R}^c$, the Cauchy--Schwarz inequality gives
$\|z\|_1 = \langle |z|, \mathbf{1}\rangle \leq \|z\|_2\,\|\mathbf{1}\|_2
= \sqrt{c}\,\|z\|_2$. Applying this:
\begin{equation}
    \|J_f(x_0)\,v_i\|_1 \;\leq\; \sqrt{c}\,\|J_f(x_0)\,v_i\|_2
\end{equation}

\paragraph{Step 3: Apply the chain rule.}

Since $f(x) = W\phi(x)$, the chain rule gives $J_f(x_0) = W J_\phi(x_0)$, so:
\begin{equation}
    \|J_f(x_0)\,v_i\|_2 \;=\; \|W\,J_\phi(x_0)\,v_i\|_2
\end{equation}

\paragraph{Step 4: Use the SVD identity $J_\phi(x_0)\,v_i = \sigma_i u_i$.}

By definition of the SVD $J_\phi(x_0) = U\Sigma V^\top$, the right singular
vectors are mapped as $J_\phi(x_0)\,v_i = \sigma_i\,u_i$
where $u_i \in \mathbb{R}^m$ is the $i$-th left singular vector with
$\|u_i\|_2 = 1$. Substituting:
\begin{equation}
    \|W\,J_\phi(x_0)\,v_i\|_2 \;=\; \|W\,\sigma_i\,u_i\|_2
    \;=\; \sigma_i\,\|W\,u_i\|_2
\end{equation}

\paragraph{Step 5: Bound by the operator norm.}

By definition, $\|W\|_{\mathrm{op}} = \max_{\|v\|=1}\|Wv\|_2$. Since
$\|u_i\|_2 = 1$:
\begin{equation}
    \|W\,u_i\|_2 \;\leq\; \|W\|_{\mathrm{op}}
\end{equation}

\paragraph{Step 6: Chain the bounds and conclude.}

Combining Steps 2--5:
\begin{equation}
    \|J_f(x_0)\,v_i\|_1
    \;\leq\; \sqrt{c}\,\sigma_i\,\|W\|_{\mathrm{op}}
\end{equation}
Substituting into Step 1:
\begin{equation}
    \mathrm{GI}(f, x_0, v_i)
    \;\geq\; \frac{q_0 - \delta}{\sqrt{c}\,\|W\|_{\mathrm{op}}\,\sigma_i}
\end{equation}
\hfill$\square$

\section{Proof of Theorem~\ref{thm:gi_sharpness}}
\label{app:proof_thm_gi_sharpness}

We prove that the input-space loss gradient is bounded by
$\sqrt{2}\|W\|_{\mathrm{op}}\cdot\sigma_1$.

\paragraph{Step 1: Express the input gradient via the chain rule.}

The cross-entropy loss is $\mathcal{L}(x) = -\log\sigma_y(f(x))$, where
$\sigma_y$ denotes the softmax probability of the true class $y$.
Differentiating with respect to the input $x$ via the chain rule:
\begin{equation}
    \nabla_x \mathcal{L}(x_0)
    \;=\; J_f(x_0)^\top \,g
\end{equation}
where $g \in \mathbb{R}^c$ is the gradient of the loss with respect to the
logits $f(x_0)$, with entries $g_i = \sigma_i(x_0) - \mathbf{1}[i = y]$.

\paragraph{Step 2: Bound $\|g\|_2$.}

Since $g_y = \sigma_y(x_0) - 1$ and $g_i = \sigma_i(x_0)$ for $i \neq y$,
with $\sigma_i(x_0) \in [0,1]$ and $\sum_i \sigma_i = 1$, the squared norm satisfies
$\|g\|_2^2 = (\sigma_y - 1)^2 + \sum_{i \neq y} \sigma_i^2 \leq (\sigma_y - 1)^2 + (1 - \sigma_y)^2 = 2(1-\sigma_y)^2 \leq 2$,
so $\|g\|_2 \leq \sqrt{2}$.

\paragraph{Step 3: Bound the input gradient norm.}

Taking the $\ell^2$ norm and applying submultiplicativity, then using the chain rule decomposition $J_f(x_0) = WJ_\phi(x_0)$ and $\|J_\phi(x_0)\|_2 = \sigma_1$:
\begin{equation}
    \|\nabla_x \mathcal{L}(x_0)\|_2
    \;\leq\; \|W\|_{\mathrm{op}} \cdot \sigma_1 \cdot \sqrt{2}
    \;=\; \sqrt{2}\,\|W\|_{\mathrm{op}} \cdot \sigma_1
\end{equation}
This establishes Equation~\eqref{eq:loss_grad_bound}. \hfill$\square_{\text{part 1}}$

\section{Proof of Theorem~\ref{thm:gi_adversarial}}
\label{app:proof_thm_gi_adversarial}

We prove two claims: (1) any successful adversarial perturbation $\Delta$
must satisfy $\|\Delta\|_2 \geq \min_{\|d\|=1}\mathrm{GI}(f,x_0,d)$, and
(2) the lower bound
$\frac{q_0 - \delta}{\sqrt{c}\,\|W\|_{\mathrm{op}}\sigma_1}$.

\paragraph{Step 1: Decompose $\Delta$ into direction and magnitude.}

Any nonzero perturbation $\Delta \in \mathbb{R}^d$ can be written as
$\Delta = \|\Delta\|_2 \cdot d_\Delta$ where $d_\Delta = \Delta/\|\Delta\|_2$
and $\|d_\Delta\| = 1$.

\paragraph{Step 2: Derive the lower bound by contradiction.}

Suppose the perturbation $\Delta$ successfully changes the classification:
$\arg\max_i f^i(x_0 + \Delta) \neq y$.
Suppose for contradiction that $\|\Delta\|_2 < \mathrm{GI}(f, x_0, d_\Delta)$.
By definition of GI, moving distance $\|\Delta\|_2$ in direction $d_\Delta$ does not change the predicted class, which contradicts the assumption that the attack succeeds. Therefore:
\begin{equation}
    \|\Delta\|_2 \;\geq\; \mathrm{GI}(f, x_0, d_\Delta)
\end{equation}

\paragraph{Step 3: Optimize over the adversary's choice of direction.}

Since the bound holds for any direction $d_\Delta$, the tightest constraint the adversary faces is:
\begin{equation}
    \|\Delta\|_2 \;\geq\; \min_{\|d\|=1}\,\mathrm{GI}(f, x_0, d)
\end{equation}
This establishes Equation~\eqref{eq:adv_radius_gi}.
\paragraph{Step 4: Apply Theorem \ref{thm: lin-appx-gi}.}
Observe that for any unit $d$, by Cauchy-Schwarz and the definition of operator norm,
\begin{equation}
    \|J_f(x_0)d\|_1 \leq \sqrt{c}\|W\|_{\mathrm{op}}\|J_{\phi}d\|_2 \leq \sqrt{c}\|W\|_{\mathrm{op}}\sigma_1
\end{equation}
Then, applying this bound to Theorem\ref{thm: lin-appx-gi}, we obtain the desired bound.

\hfill$\square$

\section{Understanding GI Growth During Training}
\label{sec:gi_growth_understanding}

Our experiments in Section~\ref{sec:gi_evolution} (and specifically Figure~\ref{fig:subfig_a}, which compares epoch 1 to epoch 50) show consistently that GIs expand during training, and that this expansion is anisotropic: GIs grow most along high-index RSV directions (small $\sigma_i$) and shrink along low-index RSV directions (large $\sigma_i$). Here we provide a unified understanding of why this happens, grounded in two mechanisms that our experiments and prior work support.

The GI lower bound from Theorem~\ref{thm:gi_spectral} is:
$$B_i(t) \;=\; \frac{q_0(t) - \delta}{\sqrt{c}\,\|W(t)\|_{\mathrm{op}}\,\sigma_i(t)}$$

This bound grows when the numerator (margin) increases and when the denominator (sensitivity) decreases. We identify two mechanisms driving this:

\noindent\textbf{(i) Margin growth.} \citet{Lyu2020Gradient} prove that for homogeneous networks trained with cross-entropy loss, a smooth approximation of the normalized margin increases monotonically during training. This supports the numerator $q_0(t)$ being non-decreasing, consistent with what we observe empirically.

\noindent\textbf{(ii) Anisotropic restructuring of Jacobian singular values.} As seen in Figure~\ref{fig:gi_expansion_and_linear_approximation}(a), training does not uniformly shrink all singular values. Instead, it \emph{sharpens the anisotropy}: high-index singular values (which are already small) shrink further, while low-index singular values (which are already large) may grow.

This observation provides a finer-grained directional decomposition of the aggregate findings by \citet{novak2018sensitivity}. They report that the overall Jacobian Frobenius norm, $\|J\|_F = \sqrt{\sum_i \sigma_i^2}$, decreases during training. Because there are many more high-index (small $\sigma_i$) directions than low-index ones, the net effect on $\|J\|_F$ is still a decrease, even while a few low-index $\sigma_i$ grow. It is also important to note a structural distinction: while \citet{novak2018sensitivity} measure the input-output Jacobian $J_f = \frac{\partial f(x)}{\partial x}$, our analysis relies on the penultimate layer Jacobian $J_\phi = \frac{\partial \phi(x)}{\partial x}$. These are related by $J_f = W J_\phi$, meaning their singular value spectra are connected but not identical. Our empirical results directly measure $J_\phi$, making them specifically relevant to our framework.

The consequence of this restructuring is that $B_i(t)$ grows fastest in directions where $\sigma_i$ is already small (high-index RSVs), because $\partial B_i / \partial \sigma_i \propto -1/\sigma_i^2$ is largest in magnitude there. This directly explains the skewed GI distribution seen in Figure~\ref{fig:gi_expansion_and_linear_approximation}(a): after training, GIs are disproportionately large along high-index RSV directions. We note that a fully rigorous proof of individual $\sigma_i$ decay during training remains an open theoretical question; our account here is grounded in our own empirical observations of $J_\phi$ and the aggregate sensitivity results of \citet{novak2018sensitivity}.

\end{document}